\documentclass{article}

\def\conference{arxiv}
\def\neuripsconf{neurips}
\def\icmlconf{icml}
\def\arxivconf{arxiv}

\usepackage[table,xcdraw]{xcolor}

\ifx\conference\neuripsconf
    \usepackage[nonatbib]{styles/neurips_2025}
    \usepackage[numbers,sort&compress]{natbib}
\else\ifx\conference\arxivconf
    \usepackage{styles/arxiv}
\else
    \usepackage{styles/icml2025}
\fi
\fi

\usepackage{enumitem}

\usepackage[utf8]{inputenc} 
\usepackage[T1]{fontenc}    
\usepackage{amsmath}
\usepackage{hyperref}       
\usepackage{url}            
\usepackage{booktabs}       
\usepackage{amsfonts}       
\usepackage{nicefrac}       
\usepackage{microtype}      
\usepackage{xspace}
\usepackage{cleveref}
\usepackage[skins,breakable]{tcolorbox}
\usepackage{listings}
\usepackage{bbold}
\usepackage{enumitem}
\usepackage{algorithm}
\usepackage{algpseudocode}
\usepackage{ifthen}
\usepackage{fontawesome5}

\usepackage{amsthm}
\usepackage{math_commands}
\usepackage{amssymb}
\usepackage{mathtools}

\usepackage{pifont}

\usepackage{graphicx}
\usepackage{makecell}
\usepackage{balance}
\usepackage{flushend}
\usepackage{soul}
\usepackage{mwe}
\usepackage{subcaption}
\usepackage{multirow}
\usepackage{colortbl}
\usepackage{wrapfig}
\tcbuselibrary{listings}

\definecolor{slabg}{HTML}{EAF1FB}
\definecolor{gainc}{HTML}{1E7B34}
\definecolor{subtle}{gray}{0.42}
\newcommand{\pmstd}[1]{{\scriptsize$\,\pm\,$#1}}
\newcommand{\second}[1]{\underline{#1}}
\newcommand{\metric}[1]{{\footnotesize\color{subtle}#1}}
\newcommand{\gaindown}[1]{{\footnotesize\color{gainc}$-$#1\%}}
\newcommand{\NR}{{\color{subtle}NR}}
\definecolor{dropc}{HTML}{B3261E}
\newcommand{\gainabs}[1]{{\footnotesize\color{gainc}$+$#1}}
\newcommand{\dropabs}[1]{{\footnotesize\color{dropc}$-$#1}}
\newcommand{\dgood}[1]{{\footnotesize\color{gainc}#1}}
\newcommand{\dbad}[1]{{\footnotesize\color{dropc}#1}}

\newcounter{packednmbr}

\newenvironment{packeditemize}{\begin{list}{$\bullet$}{\setlength{\itemsep}{0.5pt}\addtolength{\labelwidth}{-4pt}\setlength{\leftmargin}{2ex}\setlength{\listparindent}{\parindent}\setlength{\parsep}{1pt}\setlength{\topsep}{2pt}}}{\end{list}}

\definecolor{refinegreen}{RGB}{0, 128, 75}
\definecolor{scoregreen}{RGB}{34, 139, 34}

\definecolor{darkgray}{RGB}{70,70,70}
\definecolor{lightgray}{RGB}{240,240,240}

\tcbset{
    boxstyle/.style={
        enhanced,
        sharp corners,
        colback=gray!20,
        colframe=gray!60,
        boxrule=1pt,
        left=10pt,
        right=10pt,
        top=5pt,
        bottom=5pt
    },
    headerstyle/.style={
        enhanced,
        sharp corners,
        colback=gray!60,
        coltext=white,
        boxrule=0pt,
        left=10pt,
        right=10pt,
        top=5pt,
        bottom=5pt,
        fontupper=\bfseries
    },
    verbatimstyle/.style={
        enhanced,
        sharp corners,
        colback=gray!10,
        colframe=gray!40,
        boxrule=1pt,
        left=10pt,
        right=10pt,
        top=5pt,
        bottom=5pt,
        listing only,
        fontupper=\ttfamily\small
    }
}

\newtcolorbox{examplebox}[1][]{
    colback=lightgray!10,
    colframe=black,
    boxrule=0.75pt,
    title=#1,
    fonttitle=\bfseries,
    left=3pt,
    right=3pt,
    top=2pt,
    bottom=2pt,
    breakable,  
}

\newtcolorbox{exampleboxcode}[1][]{
    colback=lightgray!10,
    colframe=blue,
    boxrule=0.75pt,
    title=#1,
    fonttitle=\ttfamily,
    left=3pt,
    right=3pt,
    top=2pt,
    bottom=2pt,
}

\definecolor{codegreen}{rgb}{0,0.6,0}
\definecolor{codegray}{rgb}{0.5,0.5,0.5}
\definecolor{codepurple}{rgb}{0.58,0,0.82}
\definecolor{backcolour}{rgb}{0.97,0.97,0.97}

\lstdefinestyle{codestyle}{
    commentstyle=\color{codegreen},
    keywordstyle=\color{magenta},
    numberstyle=\tiny\color{codegray},
    stringstyle=\color{codepurple},
    basicstyle=\ttfamily\scriptsize,
    breakatwhitespace=false,         
    breaklines=true,                 
    captionpos=b,                    
    keepspaces=true,                 
    numbers=left,                    
    numbersep=5pt,                  
    showspaces=false,                
    showstringspaces=false,
    showtabs=false,                  
    tabsize=2,
    xleftmargin=10pt,                
    xrightmargin=10pt                
}
\makeatletter
\newcommand{\github}[1]{%
   \href{#1}{\faGithubSquare}%
}
\makeatother

\ifx\conference\neuripsconf

\title{\artifacts: Stateless Language Agents: Scaling Long-Horizon Automated Research}
\author{%
  David S.~Hippocampus\thanks{Use footnote for providing further information
    about author (webpage, alternative address)---\emph{not} for acknowledging
    funding agencies.} \\
  Department of Computer Science\\
  Cranberry-Lemon University\\
  Pittsburgh, PA 15213 \\
  \texttt{hippo@cs.cranberry-lemon.edu} \\
}

\else\ifx\conference\icmlconf
\icmltitlerunning{Submission and Formatting Instructions for ICML 2025}

\twocolumn[
\icmltitle{\artifacts: Stateless Language Agents: Scaling Long-Horizon Automated Research}
\icmlsetsymbol{equal}{*}

\begin{icmlauthorlist}
\icmlauthor{Firstname1 Lastname1}{equal,yyy}
\icmlauthor{Firstname2 Lastname2}{equal,yyy,comp}
\icmlauthor{Firstname3 Lastname3}{comp}
\icmlauthor{Firstname4 Lastname4}{sch}
\icmlauthor{Firstname5 Lastname5}{yyy}
\icmlauthor{Firstname6 Lastname6}{sch,yyy,comp}
\icmlauthor{Firstname7 Lastname7}{comp}
\icmlauthor{Firstname8 Lastname8}{sch}
\icmlauthor{Firstname8 Lastname8}{yyy,comp}
\end{icmlauthorlist}

\icmlaffiliation{yyy}{Department of XXX, University of YYY, Location, Country}
\icmlaffiliation{comp}{Company Name, Location, Country}
\icmlaffiliation{sch}{School of ZZZ, Institute of WWW, Location, Country}

\icmlcorrespondingauthor{Firstname1 Lastname1}{first1.last1@xxx.edu}
\icmlcorrespondingauthor{Firstname2 Lastname2}{first2.last2@www.uk}

\icmlkeywords{Machine Learning, ICML}

\else\ifx\conference\arxivconf

\title{Stateless Language Agents: Scaling Long-Horizon Automated Research}

\author{\textbf{Qizheng Zhang} \textsuperscript{1}$^*$
\quad
\textbf{Changxiu Ji} \textsuperscript{1}$^*$
\quad
\textbf{Isaac Sun} \textsuperscript{2}$^*$
\quad
\textbf{Yuetai Li} \textsuperscript{3}
\quad
\textbf{Shubhangi Upasani} \textsuperscript{4} \\
\textbf{Sherry Ruan} \textsuperscript{1}
\quad
\textbf{Boyuan Ma}  \textsuperscript{4}
\quad
\textbf{Fenglu Hong} \textsuperscript{4}
\quad
\textbf{Vamsidhar Kamanuru} \textsuperscript{4}
\quad
\textbf{Yoonho Lee} \textsuperscript{1} \\
\textbf{Yuzhen Mao} \textsuperscript{1} 
\quad
\textbf{Genghan Zhang} \textsuperscript{1} 
\quad
\textbf{Rulin Shao} \textsuperscript{3} 
\quad
\textbf{Qiuyang Mang} \textsuperscript{5} 
\quad
\textbf{Andy Dimnaku} \textsuperscript{1} \\
\textbf{Changran Hu} \textsuperscript{4} 
\quad
\textbf{Radha Poovendran} \textsuperscript{3} 
\quad
\textbf{Kunle Olukotun} \textsuperscript{1}\\[5pt]
\textsuperscript{1} Stanford University \quad \textsuperscript{2} Carnegie Mellon University \quad \textsuperscript{3} University of Washington\\ \textsuperscript{4} SambaNova Systems, Inc. \quad \textsuperscript{5} UC Berkeley \quad $^*$ equal contribution\\[3pt]
\faEnvelope \enspace \texttt{qizhengz@stanford.edu, kunle@stanford.edu} \quad
\faGithubSquare \enspace \texttt{\href{https://github.com/Alex-q-z/stateless-language-agents}{SLA}}
}

\begin{document}

\maketitle
\fi
\fi
\fi

\begin{figure}[ht]
    \centering
    \vspace{-15pt}
    \includegraphics[width=\linewidth]{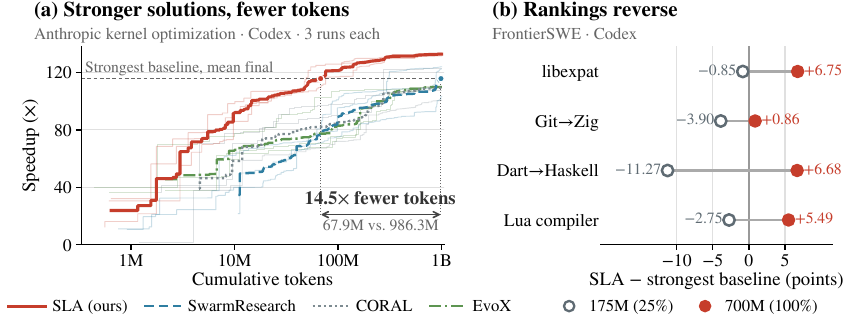}
    \caption{\textbf{SLA sustains research progress over long horizons.}
    \textbf{(a)} On Anthropic kernel optimization, SLA finds stronger solutions and reaches the strongest baseline's mean final performance with 14.5$\times$ fewer tokens (a 93.1\% reduction); every SLA run finishes ahead of every baseline run. Thick curves show means over three runs per method; thin curves show individual runs.
    \textbf{(b)} On all four FrontierSWE tasks, SLA trails the strongest per-task baseline at 175M tokens (25\% of the budget, open circles) but leads at 700M (full budget, filled circles).
    All results use Codex with GPT-5.5.}
    \label{fig:teaser}
\end{figure}
\begin{abstract}
Automated research systems increasingly run LLM agents over long horizons, but more inference does not by itself produce more progress: agents replay growing histories, duplicate one another's work, or stop experimenting while token consumption continues. Yet most evaluations use short budgets or benchmarks that saturate early, leaving these failure modes untested. We trace these failures to two choices: where research state lives and who decides what to try next. We introduce \textbf{Stateless Language Agents (SLAs)}, built on the principle of \textbf{stateful search with stateless agents}: no agent carries its conversation across invocations; instead, the harness owns the research state (candidate solutions and measured outcomes) and reconstructs a fresh and role-specific context for every invocation. What each agent sees becomes an explicit design choice rather than a history that grows with the run. We implement this principle in the SLA framework, where a stateless Advisor reads harness-summarized evidence across search directions and assigns concrete experiments to parallel Workers. We evaluate SLA against three recent frameworks on software engineering, kernel optimization, and algorithm design at budgets of up to one billion tokens. SLA achieves the best final result on every task and reaches the strongest kernel baseline's final performance with over 84\% fewer tokens. Ablations from shared checkpoints show that focused contexts and explicit assignments each contribute to SLA's progress, with effects that can compound over full runs, while the Advisor consumes less than 0.6\% of tokens. These results argue for SLAs, which keep durable research state out of agent conversations, and show that short evaluation horizons can misjudge research systems and their components.
\end{abstract} 
\section{Introduction}

Large language model (LLM) agents are increasingly used for automated research (AutoResearch): iteratively proposing, implementing, and evaluating solutions to problems with measurable objectives but no known optimum~\citep{romera2024mathematical, novikov2025alphaevolve, mang2025frontiercs, coral, swarmresearch}. As these systems take on harder problems, what matters is not only how quickly they find good solutions, but whether additional computation keeps producing better ones. In other words, \textbf{how can automated research sustain progress as its inference budget grows?}

Existing evaluations rarely run long enough to answer this question. Most measure budgets in iterations or model calls rather than tokens. Widely used benchmarks also saturate early: on 26-circle packing, four methods finish within $2.3 \times 10^{-10}$ of one another by 5M tokens (Figure~\ref{fig:challenge-motivation}(a)). A saturated benchmark measures how quickly a method reaches the ceiling, not whether it can keep converting computation into progress.
As a result, design choices whose effects compound as experience accumulates, such as context management and experiment selection, remain largely untested.

Long runs expose two challenges. First, \textit{retaining experience while controlling context}: candidate solutions, evaluation results, and failed attempts are reusable evidence, but replaying a growing history consumes tokens and degrades agent behavior~\citep{hong2025context, xia2026diagnosing}. Second, \textit{turning accumulated evidence into productive exploration}: recorded findings do not determine what to try next. In one CORAL~\citep{coral} run on a GPU kernel task, the last improvement arrives at 231M tokens. Afterward, agents repeatedly declare their solutions optimal, and 98\% of the final 1,500 sessions make no tool calls while token consumption continues (Figure~\ref{fig:challenge-motivation}(b)).
This shows that \textbf{scaling inference alone does not guarantee productive search.}

We introduce \textbf{Stateless Language Agents (SLAs)}, built on one principle: \textbf{stateful search with stateless agents}. We distinguish an agent's \textit{state}, which persists across invocations, from its \textit{context}, which is what it sees within one invocation. A stateless agent keeps no state: nothing it accumulates, including its conversation, carries over to its next invocation. This does not make it contextless. The harness owns the research state (candidate solutions and measured outcomes) and reconstructs a fresh and role-specific context from it for every invocation. What each agent sees thus becomes an explicit design choice, not a by-product of accumulated conversation.
We implement this principle in \textbf{the SLA framework}, which addresses each challenge with one mechanism. With \textit{harness-controlled context reconstruction}, a stateless Advisor sees structured evidence across all search directions, while each stateless Worker sees only its assignment, its retained candidate, and local feedback. With \textit{evidence-driven work assignment}, the Advisor turns this evidence into concrete experiments, so parallel Workers run distinct experiments instead of repeating each other's work. Harness-owned state also makes SLA an experimental instrument: our ablations resume from identical checkpointed research states.

We evaluate SLA against EvoX~\citep{evox}, CORAL~\citep{coral}, and SwarmResearch~\citep{swarmresearch} on long-horizon software engineering, kernel optimization, and algorithm design at budgets of up to one billion cumulative tokens, measuring both solution quality at matched budgets and the tokens needed to reach a common target. Our evaluation yields three findings:
\begin{packeditemize}
    \item \textbf{SLA finds stronger solutions with fewer tokens.} SLA achieves the best full-budget result on every evaluated task. On Anthropic kernel optimization with Codex, all three SLA runs outperform all nine baseline runs. SLA also reaches the strongest baseline's final performance with 93\% fewer tokens (84\% with Claude Code). On FrontierSWE, it improves scores by 4.95 points on average over the strongest per-task baseline.
    \item \textbf{Focused contexts and explicit assignments each contribute to SLA's progress, and their effects can compound over a run.} From shared checkpoints, removing Advisor context reconstruction, Worker context isolation, or Advisor assignments reduces mean progress in every comparison. Removing Worker isolation costs 5–11 cycles over 100M-token continuations but 224 cycles over a full 1B-token run. Without assignments, parallel Workers repeatedly implement the same feature (Appendix~\ref{sec:extended-trace-analysis}).
    \item \textbf{Explicit coordination is cheap.} The Advisor consumes 0.24–0.51\% of tokens and 1.2–2.3\% of model cost, compared with 8.4–10.7\% of tokens and 6.1–7.4\% of cost for SwarmResearch's Shepherd. Because coordination is separated from implementation, the Worker model can also be chosen per task (\S\ref{subsec:scaling-and-cost}).
\end{packeditemize}

Our results also have implications for evaluating AutoResearch systems. Across our experiments, short horizons can misjudge methods, components, and configurations. On FrontierSWE, SLA trails the strongest baseline at 175M tokens but leads at 700M (\S\ref{subsec:main-results}). The cost of removing Worker isolation grows by more than an order of magnitude between a 100M-token continuation and a full run (\S\ref{subsec:ablation-and-more}). Wider Worker pools that trail by 337 cycles at 25\% of the budget finish within 4 cycles of the narrowest (\S\ref{subsec:scaling-and-cost}). We therefore report results at multiple cumulative token budgets and encourage future evaluations to do the same.
\section{Background and Motivation}

\subsection{Related Work on Automated Research (AutoResearch)}
\label{subsec:background-related-work}

Automated research (AutoResearch) uses LLMs to iteratively propose, implement, and evaluate ideas for open-ended scientific and engineering problems.
It targets problems with measurable objectives but often no known optimal solution.
A major line of work, which we call LLM-guided search, embeds LLM-generated programs in evolutionary search loops and uses automated evaluation to guide improvements.
Examples include FunSearch~\citep{romera2024mathematical}, AlphaEvolve~\citep{novikov2025alphaevolve}, OpenEvolve~\citep{openevolve}, ShinkaEvolve~\citep{lange2026shinkaevolve}, EvoX~\citep{evox}, and AdaEvolve~\citep{adaevolve}, spanning mathematical optimization, algorithm design, and systems optimization.
In these methods, each candidate comes from one or a few model calls.

Recent agent-driven frameworks, including CORAL~\citep{coral} and SwarmResearch~\citep{swarmresearch}, place more of the experimental process under the control of tool-using coding agents such as OpenAI Codex~\citep{openaicodex} and Claude Code~\citep{claudecode}.
Agents can inspect code, implement ideas, run tests, and revise their approach over many steps, while multiple agents explore different directions.
In these frameworks, agents also carry much of the research history in their own long-running sessions.
The shift to coding agents motivates studying how such research processes sustain progress and use computation efficiently over long horizons.

Other work studies broader AI scientist systems that automate or assist multiple stages of scientific discovery, including hypothesis generation, literature review, experimentation, and scientific writing~\citep{lu2024ai,yamada2025ai,gottweis2025towards}.
We instead focus on open-ended search over executable solutions with automated evaluators.
This setting lets us isolate how agentic research processes sustain progress as the search horizon grows.
We discuss related work further in Appendix~\ref{app:related-work}.

\subsection{Long-Horizon Evaluation and Its Challenges}
\label{subsec:short-horizon-limitation}
\label{subsec:challenges}

\begin{figure}[t]
    \centering
    \includegraphics[width=\linewidth]{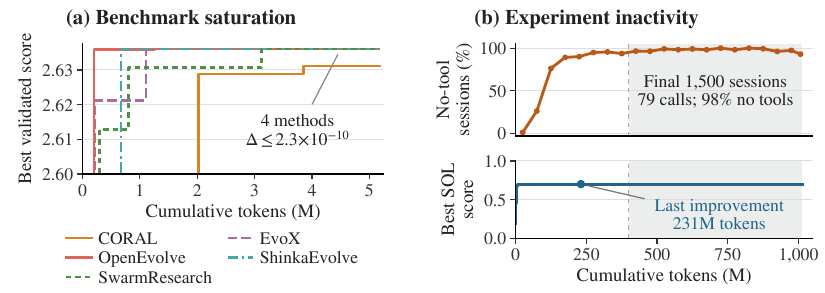}
    \caption{\textbf{Long-horizon AutoResearch requires benchmark headroom and sustained experimentation.}
    \textbf{(a)} On 26-circle packing, four methods reach nearly identical scores by 5M tokens (within $2.3 \times 10^{-10}$), leaving little separation between methods.
    \textbf{(b)} In a CORAL run on a GPU kernel task, the last improvement occurs at 231M tokens, but token consumption continues to 1B. Of the final 1,500 sessions, 98\% make no tool calls, showing that additional inference does not always translate to productive experiments.}
    \vspace{-10pt}
    \label{fig:challenge-motivation}
\end{figure}

\paragraph{Measuring the research horizon.}
Most AutoResearch evaluations define budgets by iterations, model calls, or evaluator queries. CORAL~\citep{coral} runs agents over long wall-clock periods but does not report model tokens. Wall-clock time mixes model latency, evaluator runtime, and system overhead. It also depends on parallelism: under the same wall-clock budget, a multi-agent system can consume more inference than a single agent. We therefore measure the horizon in cumulative input and output tokens (\emph{cumulative tokens}). With fixed models and inference settings, cumulative tokens give a controlled and reproducible scaling axis. Long horizons also require benchmarks with headroom. On 26-circle packing, popularized by AlphaEvolve~\citep{novikov2025alphaevolve}, two methods come within $10^{-3}$ of the best observed score by 0.67M tokens. After that point, the benchmark mainly measures how quickly methods reach the ceiling (Figure~\ref{fig:challenge-motivation}(a)).

\paragraph{Retaining experience while controlling context.}
Candidate solutions, evaluation results, and failed attempts may become relevant only later. In our CORAL FrontierSWE Lua trace, an agent revisits an older experimental note and improves a later leading candidate by adding a missing formatting feature (Appendix~\ref{app:coral-lua-trace}). Yet repeatedly processing a growing history consumes tokens, and accumulated context can degrade agent behavior~\citep{hong2025context}, including premature stopping in long-horizon search~\citep{xia2026diagnosing}.
In recent work, Liu et al.~\citep{liu2026agents} compare coding-agent sessions of up to 100M tokens with an independent-sampling reference. Agents first improve faster than sampling but fall below it as sessions grow, so splitting a fixed budget across several shorter sessions can outperform one long session.
The challenge is to keep reusable evidence while selecting what each agent needs for its current decision.

\paragraph{Turning accumulated evidence into productive exploration.}
Recorded findings do not determine what to try next.
In our CORAL FrontierSWE libexpat trace, two agents implement the same feature from the same parent and get the same score. They then overlap again on the next feature, even though one agent had posted a note claiming it (Appendix~\ref{app:coral-libexpat-trace}).
In the GPU kernel run of Figure~\ref{fig:challenge-motivation}(b), agents repeatedly declare their candidates optimal and wait, despite prompts to continue and restarts. CORAL resumes each agent's own session after a restart. A restarted agent therefore continues from the same conversation in which it decided to stop, and additional inference largely repeats that decision. The challenge is to turn accumulated outcomes into concrete next experiments, coordinate parallel effort, and revise directions as evidence arrives.
\section{The SLA (Stateless Language Agents) Framework}

\begin{figure}[t]
    \centering
    \includegraphics[width=\linewidth]{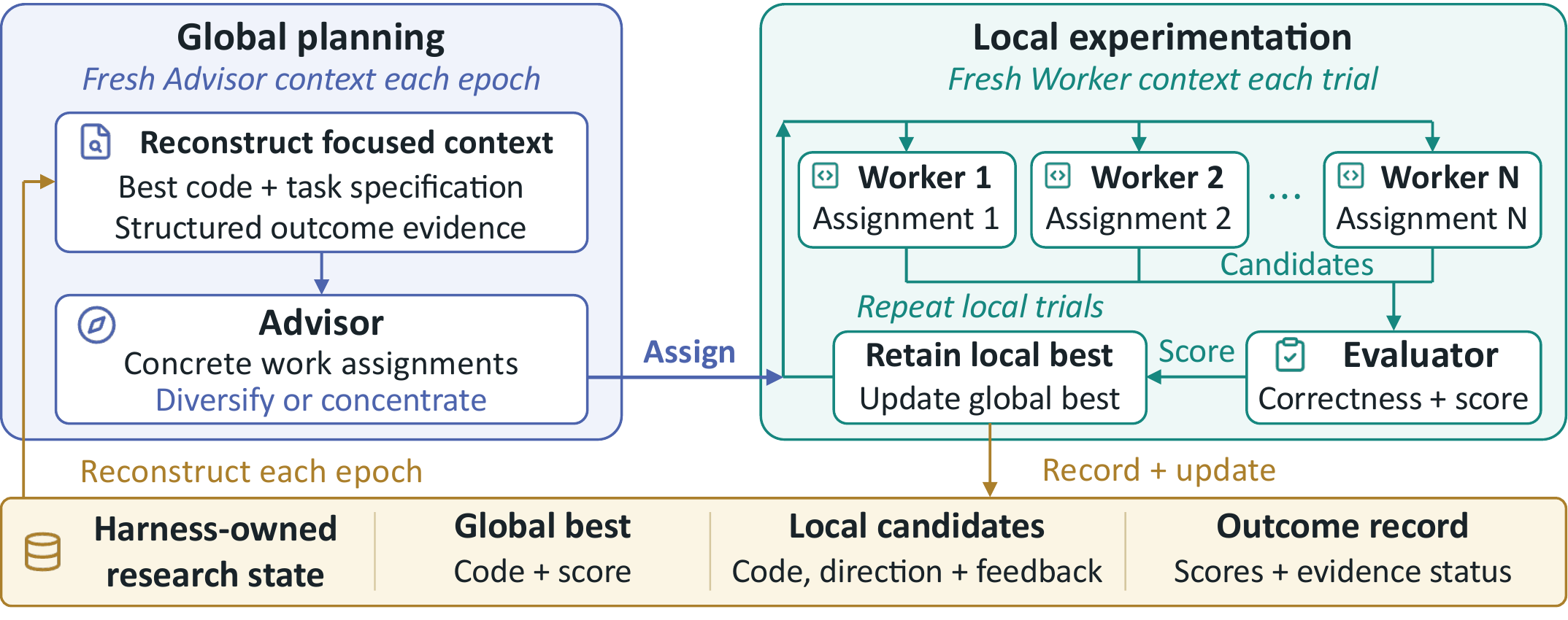}
    \caption{\textbf{The SLA framework: stateful search with stateless agents.}
    The harness owns candidates and outcomes and rebuilds a fresh and role-specific context for the Advisor each epoch and for each Worker each trial (\S\ref{sec:design-context}). The Advisor turns global evidence into coordinated assignments (\S\ref{sec:design-steering}).}
    \label{fig:workflow}
\end{figure}

The SLA framework applies stateless language agents to long-horizon automated research (Figure~\ref{fig:workflow}). It alternates global planning with local experimentation. Each epoch, the harness reconstructs the Advisor's context from the research state (\S\ref{sec:design-context}), and the Advisor assigns concrete experiments to parallel Workers (\S\ref{sec:design-steering}). Each Worker runs several local trials in an isolated workspace. An independent evaluator scores every candidate. The harness records outcomes and updates each Worker's local candidate and the global best until the budget is exhausted. Each Advisor epoch and each Worker trial is one invocation. Every invocation starts a new agent session, and only what the harness records in the research state carries over to the next one.

\paragraph{Relation to native subagent orchestration.}
Coding agents increasingly support native subagents, which run in parallel with separate contexts. The orchestrating agent, however, remains stateful: it keeps its own conversation and decides what each subagent sees and when to stop. SLA is compatible with such tools, but every SLA agent, including the Advisor, is stateless, and the harness decides what each context contains.

\subsection{Harness-Controlled Context Reconstruction}
\label{sec:design-context}

SLA separates state, which the harness retains, from context, which each agent sees. The Advisor needs evidence across search directions to choose experiments. Each Worker needs only the local information required to develop its assignment. The harness therefore rebuilds both contexts from the research state at every Advisor epoch and Worker trial.

\textbf{Advisor context: evidence for global decisions.}
Each epoch, the Advisor receives the task specification, the current global best code and score, and an evidence summary built by the harness. The summary covers recent attempts, outcomes grouped by search direction, Worker status, and recent progress, plus task-specific diagnostics when available. It separates measured non-improvements from inconclusive failures in implementation, runtime, or correctness, so a failed execution is not mistaken for evidence that a direction is exhausted. The harness keeps a full chronological record of every experiment, but the Advisor sees only this reconstructed context. Like PEEK's context map~\citep{gu2026peek}, this context lets the Advisor compare directions without reading the full chronological record.

\textbf{Worker context: evidence for focused experimentation.}
Each trial, a Worker receives the task specification, its assignment, its retained candidate, and compact local feedback: the previous trial's status, measured score, and available correctness diagnostics. Workers cannot access other Workers' workspaces or the full research history. Evidence from other directions reaches them only through the Advisor's assignments. Because candidate code and feedback persist in the research state, Workers build on prior progress without inheriting the conversations that produced it.

\subsection{Evidence-Driven Work Assignment}
\label{sec:design-steering}

Workers acting only on local information may repeat attempts made elsewhere or overlook useful findings. SLA therefore gives experiment selection to the Advisor, which sees evidence across all search directions and turns it into concrete, coordinated assignments.

Because the Advisor is itself stateless, it revisits these assignments each epoch from current evidence rather than from its own earlier reasoning. It can spread Workers across directions or concentrate them on distinct approaches to a promising bottleneck. Workers decide how to implement and refine an assignment over several local trials. The harness keeps improvements to each Worker's local candidate even when that candidate trails the global best. This gives alternative approaches time to mature. When the Advisor redirects a Worker, the Worker restarts from the global best. Its first successfully evaluated candidate becomes a new local baseline, so a new direction can develop even if it starts out worse.

\subsection{Implementation}
\label{subsec:implementation}

SLA uses OpenAI Codex~\citep{openaicodex} or Claude Code~\citep{claudecode} for both the Advisor and the Workers. Statelessness and isolation are enforced rather than requested. The Advisor starts each epoch in a new session. Workers run as unprivileged processes under Linux Landlock filesystem restrictions, with restricted network access, and the harness clears their saved sessions between trials. Candidates are evaluated outside Worker environments with protected evaluator code. The harness also records token usage by role and checkpoints the full research state for recovery and controlled ablations (Appendix~\ref{app:implementation}).
\section{Results}
\label{sec:results}

Our evaluation of SLA yields four main findings:
\begin{packeditemize}
    \item \textbf{SLA finds stronger solutions with fewer tokens.} SLA achieves the best full-budget result on every evaluated task. On Anthropic kernel optimization, it reaches the strongest baseline's final performance with over 84\% fewer tokens under both coding-agent configurations (\S\ref{subsec:main-results}).
    \item \textbf{Method rankings depend on the evaluation horizon.} SLA trails the strongest baseline on all four FrontierSWE tasks at one-quarter of the token budget, yet leads on all four at the full budget (\S\ref{subsec:main-results}).
    \item \textbf{Focused contexts and explicit assignments each contribute to SLA's progress, and the Advisor costs little.} From shared checkpoints, removing Advisor context reconstruction, Worker context isolation, or Advisor assignments reduces mean progress in every comparison. The cost of removing Worker isolation grows from 5–11 cycles over 100M-token continuations to 224 cycles over a full run. The Advisor accounts for less than 0.6\% of tokens and at most 2.3\% of model cost in the analyzed runs (\S\ref{subsec:ablation-and-more} and \S\ref{subsec:scaling-and-cost}).
    \item \textbf{Scaling choices are task-dependent.} Going from 1 to 15 Workers cuts elapsed time by 6.6–7.3$\times$, and wider pools that trail early can catch up over the full budget. However, the best width varies across tasks. Cheaper Workers can outperform stronger ones at matched cost on some tasks but not others (\S\ref{subsec:scaling-and-cost}).
\end{packeditemize}

\subsection{Experiment Setup}

\paragraph{Tasks and datasets.} 
We evaluate on three task categories: 
(1) long-horizon software engineering using FrontierSWE~\citep{frontierswe}; 
(2) compute kernel optimization using Anthropic's VLIW SIMD kernel optimization task~\citep{anthropic_takehome} and NVIDIA's SOL-ExecBench GPU kernel optimization tasks~\citep{sol-execbench}; 
and (3) algorithm design using FrontierCS~\citep{mang2025frontiercs}. 
From FrontierSWE, we select four problems that do not require GPU access: libexpat to x86-64 Assembly, Git to Zig, Dart to Haskell, and Lua Native Compiler.
From SOL-ExecBench, we select the three problems with the most submissions as of August 1, 2026: \#1 (the backward pass through attention softmax, dropout, and value matrix multiplication), \#58 (mixture-of-experts token sorting with prefix sums), and \#210 (fused residual addition and RMS normalization).
From FrontierCS, we select Structured-LWE, a cryptographic task requiring the recovery of valid secret vectors for public structured learning-with-errors instances.
Unlike prior evaluations centered on benchmarks like circle packing and expert placement load balancing~\citep{evox,coral,swarmresearch}, we focus on tasks with headroom at our budgets (\S\ref{subsec:challenges}).

\paragraph{Language models and coding agents.}
We evaluate two coding-agent configurations: OpenAI Codex~\citep{openaicodex} CLI v0.152.1 with GPT-5.5, and Claude Code~\citep{claudecode} v2.1.258 with Claude Opus 4.8. In our main experiments, the Advisor and Workers use the same LLM, so gains cannot come from a stronger Advisor passing knowledge to weaker Workers. Appendix~\ref{app:worker-model} studies a stronger Advisor paired with weaker Workers. In all experiments, we set the reasoning effort to high and keep each agent's default automatic context-compaction settings. In SLA, compaction can act only within one invocation, since no session carries over to the next.

\paragraph{Baselines.}
We compare against three recent AutoResearch frameworks. Each runs from a pinned upstream release with unmodified search logic (Appendix~\ref{app:baselines}). EvoX~\citep{evox} jointly evolves candidate solutions and search strategies. Because it was designed around direct LLM calls, we launch a fresh coding agent for each generation. CORAL~\citep{coral} runs parallel agents that select their own experiments and share discoveries through persistent memory and heartbeat-triggered reflection. Each CORAL agent keeps one long-running session, which CORAL resumes after every heartbeat, stall, or restart. We use the official implementation with its upstream default configuration. SwarmResearch~\citep{swarmresearch} uses a Shepherd agent that coordinates Search Agents through parent selection, agent-type selection, and selective context sharing, but is instructed not to assign specific ideas. The Shepherd runs as a long-running session, and each Optimizer starts by resuming its parent agent's conversation. We use the authors' official implementation. All methods share the same coding agents, models, seed solutions, public task materials, and protected evaluators, and the same external cumulative-token counter stops every run.

\paragraph{Replication and uncertainty.}
We evaluate each method on Anthropic kernel optimization with Codex over three independent runs and report means $\pm$ sample standard deviations (Table~\ref{tab:anthropic-main-results}). The checkpoint-continuation ablations also use three independent runs per configuration and checkpoint (Table~\ref{tab:aes-ablations}). Other main-result configurations use one run per method. A single long-horizon run can cost hundreds to thousands of dollars, so our budget does not allow replicating every task and agent configuration.

\subsection{Main Results: Performance and Token Efficiency}
\label{subsec:main-results}

Tables~\ref{tab:anthropic-main-results}--\ref{tab:sol-execbench-main-results} report best-so-far performance at 25\% and 100\% of each task's token budget, and the cumulative tokens each method needs to first match the strongest baseline's full-budget result for the same task and agent (NR: not reached). 

\paragraph{SLA finds stronger solutions at matched token budgets.} SLA achieves the best full-budget performance on every evaluated task. Relative to the strongest baseline for each task and coding-agent configuration, it reduces Anthropic kernel cycles by 8.1–12.8\% and improves FrontierSWE scores by 4.95 points on average. It also increases SOL-ExecBench scores by 5.5\% on average across the three tasks and improves the FrontierCS Structured-LWE score by 1.5 points. Gains are smallest on SOL-ExecBench \#210, where no method, including SLA, improves by more than 0.02 between 25\% and 100\% of the budget. For this task, the best kernels from all methods use Triton. Fused RMS normalization depends on a fast row-wise reduction, and Triton gives less direct control over warp-level reductions than CUDA, which may limit what any method can reach.

\paragraph{The kernel advantage persists across independent runs.} On Anthropic kernel optimization with Codex, all three SLA runs outperform all nine baseline runs at 1B tokens: even the worst SLA run beats the best baseline run (1122 versus 1191 cycles). SLA achieves $1112.0 \pm 8.9$ cycles, compared with $1275.7 \pm 133.9$ for SwarmResearch, the strongest baseline by mean performance.

\paragraph{SLA reaches strong baseline performance with fewer tokens.} The same numerical gain can take very different amounts of search, especially when the solution is already strong. For example, reducing kernel cycles from 1150 to 1100 can be harder than reducing them from 1200 to 1150. We therefore also report the tokens each method needs to reach a common target. SLA reaches this target before every baseline in all but one evaluated configuration. On the Anthropic kernel task with Codex, SLA needs only 67.9M tokens, compared with 986.3M for SwarmResearch, a 93.1\% reduction. The savings reach 95.9\% on SOL-ExecBench \#58. The exception is SOL-ExecBench \#1, where EvoX reaches the target sooner but SLA ends with the higher score.

\paragraph{Method rankings change with the evaluation horizon.} On FrontierSWE, SLA trails the strongest baseline on all four tasks at 25\% of the budget but leads on all four at the full budget (Table~\ref{tab:frontierswe-main-results}). Short-budget evaluations would therefore miss its eventual advantage across all four tasks. These reversals motivate evaluating search methods across multiple cumulative token budgets.

\begin{table*}[t]
  \centering
  \caption{\textbf{Anthropic VLIW SIMD kernel performance and token efficiency.}
    Cycles at 25\% (250M) and 100\% (1B) of the token budget, and cumulative tokens needed to first
    match the strongest baseline's full-budget cycles (NR: not reached).
    Codex entries are mean $\pm$ std over three runs (token counts use the mean curve);
    Claude Code uses one run per method.
    \textbf{Bold}: best per column; \underline{underline}: second best.
    The last row gives SLA's relative reduction over the strongest baseline in each column.}
  \label{tab:anthropic-main-results}
  \small
  \setlength{\tabcolsep}{7pt}
  \renewcommand{\arraystretch}{1.12}
  \begin{tabular}{lcccccc}
    \toprule
    & \multicolumn{3}{c}{Codex \metric{(3 runs)}}
    & \multicolumn{3}{c}{Claude Code \metric{(1 run)}} \\
    \cmidrule(lr){2-4}\cmidrule(lr){5-7}
    Method
    & \makecell{250M\\\metric{Cycles $\downarrow$}}
    & \makecell{1B\\\metric{Cycles $\downarrow$}}
    & \makecell{To match\\\metric{Tokens (M) $\downarrow$}}
    & \makecell{250M\\\metric{Cycles $\downarrow$}}
    & \makecell{1B\\\metric{Cycles $\downarrow$}}
    & \makecell{To match\\\metric{Tokens (M) $\downarrow$}} \\
    \midrule
    EvoX
    & \second{1478.3}\pmstd{112.0} & 1343.3\pmstd{4.0} & \NR
    & \second{1174} & 1153 & \NR \\
    CORAL
    & 1524.0\pmstd{101.7} & 1350.0\pmstd{134.2} & \NR
    & 1205 & \second{1130} & \second{888.3} \\
    SwarmResearch
    & 1501.7\pmstd{185.7} & \second{1275.7}\pmstd{133.9} & \second{986.3}
    & 1648 & 1243 & \NR \\
    \midrule
    \rowcolor{slabg}
    SLA (ours)
    & \textbf{1149.3}\pmstd{18.9} & \textbf{1112.0}\pmstd{8.9} & \textbf{67.9}
    & \textbf{1079} & \textbf{1039} & \textbf{138.5} \\
    \metric{\itshape \quad Reduction vs.\ best baseline}
    & \gaindown{22.3} & \gaindown{12.8} & \gaindown{93.1}
    & \gaindown{8.1}  & \gaindown{8.1}  & \gaindown{84.4} \\
    \bottomrule
  \end{tabular}
\end{table*}
\begin{table*}[t]
  \centering
  \caption{\textbf{FrontierSWE performance and token efficiency with Codex.}
    Official scores ($\times 100$) at 25\% (175M) and 100\% (700M) of the token budget, and cumulative
    tokens needed to first match the strongest baseline's full-budget score (NR: not reached).
    One run per method. \textbf{Bold}: best per column; \underline{underline}: second best.
    The last row of each panel gives SLA's score difference from the strongest baseline (points)
    and its relative token reduction: SLA trails on all four tasks at 175M but leads at 700M.}
  \label{tab:frontierswe-main-results}
  \small
  \setlength{\tabcolsep}{7pt}
  \renewcommand{\arraystretch}{1.12}
  \begin{tabular}{lcccccc}
    \toprule
    & \multicolumn{3}{c}{libexpat $\rightarrow$ x86-64 Assembly}
    & \multicolumn{3}{c}{Git $\rightarrow$ Zig} \\
    \cmidrule(lr){2-4}\cmidrule(lr){5-7}
    Method
    & \makecell{175M\\\metric{Score $\uparrow$}}
    & \makecell{700M\\\metric{Score $\uparrow$}}
    & \makecell{To match\\\metric{Tokens (M) $\downarrow$}}
    & \makecell{175M\\\metric{Score $\uparrow$}}
    & \makecell{700M\\\metric{Score $\uparrow$}}
    & \makecell{To match\\\metric{Tokens (M) $\downarrow$}} \\
    \midrule
    EvoX
    & 10.71 & 11.33 & \NR
    & 19.29 & 21.42 & \NR \\
    CORAL
    & 8.74 & 12.08 & \NR
    & 19.46 & 22.55 & \NR \\
    SwarmResearch
    & \textbf{19.67} & \second{24.58} & \second{559.9}
    & \textbf{24.32} & \second{27.98} & \second{697.3} \\
    \midrule
    \rowcolor{slabg}
    SLA (ours)
    & \second{18.82} & \textbf{31.33} & \textbf{354.1}
    & \second{20.42} & \textbf{28.84} & \textbf{630.2} \\
    \metric{\itshape \quad $\Delta$ vs.\ best baseline}
    & \dropabs{0.85} & \gainabs{6.75} & \gaindown{36.8}
    & \dropabs{3.90} & \gainabs{0.86} & \gaindown{9.6} \\
    \bottomrule
    \noalign{\vskip 7pt}
    \toprule
    & \multicolumn{3}{c}{Dart $\rightarrow$ Haskell}
    & \multicolumn{3}{c}{Lua Native Compiler} \\
    \cmidrule(lr){2-4}\cmidrule(lr){5-7}
    Method
    & \makecell{175M\\\metric{Score $\uparrow$}}
    & \makecell{700M\\\metric{Score $\uparrow$}}
    & \makecell{To match\\\metric{Tokens (M) $\downarrow$}}
    & \makecell{175M\\\metric{Score $\uparrow$}}
    & \makecell{700M\\\metric{Score $\uparrow$}}
    & \makecell{To match\\\metric{Tokens (M) $\downarrow$}} \\
    \midrule
    EvoX
    & 9.72 & 10.66 & \NR
    & 25.82 & 65.38 & \NR \\
    CORAL
    & \second{18.55} & 23.26 & \NR
    & \textbf{82.97} & \second{90.66} & \second{682.3} \\
    SwarmResearch
    & \textbf{26.70} & \second{38.59} & \second{684.1}
    & 50.55 & 65.38 & \NR \\
    \midrule
    \rowcolor{slabg}
    SLA (ours)
    & 15.43 & \textbf{45.27} & \textbf{263.6}
    & \second{80.22} & \textbf{96.15} & \textbf{354.6} \\
    \metric{\itshape \quad $\Delta$ vs.\ best baseline}
    & \dropabs{11.27} & \gainabs{6.68} & \gaindown{61.5}
    & \dropabs{2.75}  & \gainabs{5.49} & \gaindown{48.0} \\
    \bottomrule
  \end{tabular}
\end{table*}
\begin{table*}[t]
  \centering
  \caption{\textbf{SOL-ExecBench and FrontierCS performance and token efficiency with Codex.}
    SOL scores ($[0,1]$) and FrontierCS task scores at 25\% and 100\% of each task's token budget,
    and cumulative tokens needed to first match the strongest baseline's full-budget score
    (NR: not reached). One run per method.
    \textbf{Bold}: best per column; \underline{underline}: second best.
    The last row of each panel gives SLA's relative difference from the strongest baseline.}
  \label{tab:sol-execbench-main-results}
  \small
  \setlength{\tabcolsep}{7pt}
  \renewcommand{\arraystretch}{1.12}
  \begin{tabular}{lcccccc}
    \toprule
    & \multicolumn{3}{c}{SOL-ExecBench \#1}
    & \multicolumn{3}{c}{SOL-ExecBench \#58} \\
    \cmidrule(lr){2-4}\cmidrule(lr){5-7}
    Method
    & \makecell{125M\\\metric{SOL $\uparrow$}}
    & \makecell{500M\\\metric{SOL $\uparrow$}}
    & \makecell{To match\\\metric{Tokens (M) $\downarrow$}}
    & \makecell{125M\\\metric{SOL $\uparrow$}}
    & \makecell{500M\\\metric{SOL $\uparrow$}}
    & \makecell{To match\\\metric{Tokens (M) $\downarrow$}} \\
    \midrule
    EvoX
    & \textbf{0.710} & \second{0.710} & \textbf{46.1}
    & \second{0.773} & \second{0.804} & \second{354.1} \\
    CORAL
    & 0.386 & 0.552 & \NR
    & 0.694 & 0.714 & \NR \\
    SwarmResearch
    & 0.598 & 0.670 & \NR
    & 0.764 & 0.764 & \NR \\
    \midrule
    \rowcolor{slabg}
    SLA (ours)
    & \second{0.706} & \textbf{0.763} & \second{152.7}
    & \textbf{0.836} & \textbf{0.862} & \textbf{14.5} \\
    \metric{\itshape \quad $\Delta$ vs.\ best baseline}
    & \dropabs{0.6\%} & \gainabs{7.5\%} & {\footnotesize\color{dropc}$+$231.2\%}
    & \gainabs{8.2\%} & \gainabs{7.2\%} & \gaindown{95.9} \\
    \bottomrule
    \noalign{\vskip 7pt}
    \toprule
    & \multicolumn{3}{c}{SOL-ExecBench \#210}
    & \multicolumn{3}{c}{FrontierCS: Structured-LWE} \\
    \cmidrule(lr){2-4}\cmidrule(lr){5-7}
    Method
    & \makecell{125M\\\metric{SOL $\uparrow$}}
    & \makecell{500M\\\metric{SOL $\uparrow$}}
    & \makecell{To match\\\metric{Tokens (M) $\downarrow$}}
    & \makecell{50M\\\metric{Score $\uparrow$}}
    & \makecell{200M\\\metric{Score $\uparrow$}}
    & \makecell{To match\\\metric{Tokens (M) $\downarrow$}} \\
    \midrule
    EvoX
    & \second{0.375} & \second{0.386} & \second{283.0}
    & 34.0 & 55.0 & \NR \\
    CORAL
    & 0.359 & 0.375 & \NR
    & 23.5 & 38.0 & \NR \\
    SwarmResearch
    & \textbf{0.377} & 0.378 & \NR
    & \second{44.0} & \second{58.0} & \second{197.0} \\
    \midrule
    \rowcolor{slabg}
    SLA (ours)
    & \textbf{0.377} & \textbf{0.393} & \textbf{208.3}
    & \textbf{48.5} & \textbf{59.5} & \textbf{169.6} \\
    \metric{\itshape \quad $\Delta$ vs.\ best baseline}
    & \metric{0.0\%} & \gainabs{1.8\%} & \gaindown{26.4}
    & \gainabs{10.2\%} & \gainabs{2.6\%} & \gaindown{13.9} \\
    \bottomrule
  \end{tabular}
\end{table*}

\subsection{Ablation Studies: Understanding the Design Choices}
\label{subsec:ablation-and-more}

To assess the contribution of SLA's design choices, we compare continued progress from shared SLA checkpoints taken at 10\% and 20\% of each task's full token budget. Each checkpoint preserves the research state, including accumulated evidence, the global best solution, and local candidates. From each checkpoint, we continue full SLA and three variants that remove Advisor context reconstruction, Worker context isolation, or Advisor work assignments. All variants keep every agent stateless. They change only what enters each context, or whether Workers receive assignments.

Without Advisor context reconstruction, the harness gives the Advisor no computed summary. Instead, the Advisor reads the full chronological record through a bounded, read-only interface and must find and interpret the relevant evidence itself. Without Worker context isolation, Workers can read the full chronological record and every other Worker's current candidate before each trial. Without Advisor work assignments, Workers receive a prompt for an independent trial instead of the Advisor's proposal. The Advisor still runs and can still reset a Worker's candidate to the global best. Each configuration receives an additional 100M tokens per run, with three independent runs per checkpoint. Table~\ref{tab:aes-ablations} reports mean best-so-far performance after continuation.

We also run SLA without Worker context isolation for a full 1B-token run on Anthropic kernel optimization with Codex. It ends at 1336 cycles, 224 cycles worse than the mean of the three full-SLA runs (1112.0 cycles; Table~\ref{tab:anthropic-main-results}). This gap is far larger than the differences after 100M-token continuations (Table~\ref{tab:aes-ablations}), which suggests that the cost of unfocused Worker contexts compounds over longer horizons.

\paragraph{Focused contexts support research continuity.}
The two context variants keep the same research state but change what enters each context (\S\ref{sec:design-context}). Removing either Advisor context reconstruction or Worker context isolation reduces mean progress on both tasks and at both checkpoints. Keeping all evidence is therefore not enough. What each role sees from that evidence also matters.

\paragraph{Explicit assignments effectively turn evidence into progress.}
Removing Advisor assignments reduces mean progress in every comparison, even though every run starts from the same research state. This supports the division of work in \S\ref{sec:design-steering}: the Advisor turns global evidence into concrete experiments, and Workers focus on implementation and local refinement. Trace examples in Appendix~\ref{sec:extended-trace-analysis} show Workers duplicating each other's work without assignments, and the Advisor guiding one Worker to build on another's improvement in full SLA.

\begin{table*}[t]
  \centering
  \caption{\textbf{Ablations from shared SLA checkpoints.}
    In each column, every configuration resumes from the same checkpoint
    (header: start $\rightarrow$ end cumulative tokens) and runs for 100M additional tokens;
    entries are mean $\pm$ std over three continuations.
    Grey values give full SLA's progress from the checkpoint; red values give each variant's
    relative drop in that progress. \textbf{Bold}: best per column.}
  \label{tab:aes-ablations}
  \small
  \setlength{\tabcolsep}{5pt}
  \renewcommand{\arraystretch}{1.12}
  \begin{tabular}{l r@{\hspace{5pt}}l r@{\hspace{5pt}}l r@{\hspace{5pt}}l r@{\hspace{5pt}}l}
    \toprule
    & \multicolumn{4}{c}{Anthropic Kernel}
    & \multicolumn{4}{c}{FrontierSWE (libexpat)} \\
    \cmidrule(lr){2-5}\cmidrule(lr){6-9}
    Configuration
    & \multicolumn{2}{c}{\makecell{100M$\rightarrow$200M\\\metric{Cycles $\downarrow$}}}
    & \multicolumn{2}{c}{\makecell{200M$\rightarrow$300M\\\metric{Cycles $\downarrow$}}}
    & \multicolumn{2}{c}{\makecell{70M$\rightarrow$170M\\\metric{Score ($\times$100) $\uparrow$}}}
    & \multicolumn{2}{c}{\makecell{140M$\rightarrow$240M\\\metric{Score ($\times$100) $\uparrow$}}} \\
    \midrule
    {\color{subtle}\itshape Shared checkpoint}
    & {\color{subtle}1218} & & {\color{subtle}1197} &
    & {\color{subtle}16.12} & & {\color{subtle}17.69} & \\
    \midrule
    \rowcolor{slabg}
    SLA (full)
    & \textbf{1192.3}\pmstd{2.5}  & \metric{$-$25.7}
    & \textbf{1170.0}\pmstd{14.2} & \metric{$-$27.0}
    & \textbf{18.78}\pmstd{0.92}  & \metric{$+$2.66}
    & \textbf{20.13}\pmstd{0.20}  & \metric{$+$2.44} \\
    \midrule
    \multicolumn{9}{l}{\metric{\itshape Focused contexts}} \\
    \quad w/o Advisor reconstruction
    & 1198.3\pmstd{2.3} & \dropabs{23.3\%}
    & 1188.7\pmstd{6.8} & \dropabs{69.3\%}
    & 16.86\pmstd{0.64} & \dropabs{72.2\%}
    & 19.07\pmstd{0.31} & \dropabs{43.4\%} \\
    \quad w/o Worker isolation
    & 1203.3\pmstd{7.1} & \dropabs{42.8\%}
    & 1174.7\pmstd{4.5} & \dropabs{17.4\%}
    & 16.79\pmstd{0.47} & \dropabs{74.8\%}
    & 19.77\pmstd{0.12} & \dropabs{14.8\%} \\
    \addlinespace[2pt]
    \multicolumn{9}{l}{\metric{\itshape Explicit assignments}} \\
    \quad w/o Advisor assignments
    & 1195.0\pmstd{2.6} & \dropabs{10.5\%}
    & 1182.3\pmstd{8.1} & \dropabs{45.6\%}
    & 16.54\pmstd{0.33} & \dropabs{84.2\%}
    & 18.85\pmstd{0.12} & \dropabs{52.5\%} \\
    \bottomrule
  \end{tabular}
\end{table*}

\subsection{Scaling and Coordination Cost}
\label{subsec:scaling-and-cost}

\paragraph{Coordination is cheap.}
SLA spends over 99\% of its tokens on Worker search. The Advisor consumes only 0.24–0.51\% of total tokens, compared with 8.39–10.70\% for SwarmResearch's Shepherd (Table~\ref{tab:aes-token-breakdown}). Much of the Shepherd's usage comes from its long-running context. In the Anthropic kernel run, its average call reads about 99K input tokens, compared with about 8K at the start of a fresh session. Because cached inputs make up over 90\% of tokens in both methods, token share understates the cost of roles with limited cache reuse, such as the Advisor. Even so, the Advisor accounts for only 1.20–2.29\% of total model cost, compared with 6.14–7.35\% for the Shepherd. Together with the ablations in \S\ref{subsec:ablation-and-more}, this shows that explicit coordination improves progress at little overhead.

\begin{table*}[t]
  \centering
  \caption{\textbf{Token usage and coordination overhead.}
    Token composition and coordinator share (SLA Advisor or SwarmResearch Shepherd) of tokens
    and estimated cost at the full budget, using GPT-5.5 rates of \$5/\$0.50/\$30 per million
    uncached-input, cached-input, and output tokens~\citep{openai2026pricing}.
    \textbf{Bold}: lower share.}
  \label{tab:aes-token-breakdown}
  \small
  \setlength{\tabcolsep}{8pt}
  \renewcommand{\arraystretch}{1.12}
  \begin{tabular}{lccccc}
    \toprule
    & \multicolumn{3}{c}{Token composition \metric{(\%)}}
    & \multicolumn{2}{c}{Coordinator share \metric{(\%)}} \\
    \cmidrule(lr){2-4}\cmidrule(lr){5-6}
    Method
    & Cached input & Uncached input & Output
    & Tokens $\downarrow$ & Cost $\downarrow$ \\
    \midrule
    \multicolumn{6}{l}{\metric{\itshape Anthropic kernel (1B tokens)}} \\
    \quad SwarmResearch
    & 94.67 & 4.36 & 0.97 & 8.39 & 6.14 \\
    \rowcolor{slabg}
    \quad SLA (ours)
    & 93.11 & 5.59 & 1.30 & \textbf{0.24} & \textbf{1.20} \\
    \addlinespace[2pt]
    \multicolumn{6}{l}{\metric{\itshape FrontierSWE libexpat (700M tokens)}} \\
    \quad SwarmResearch
    & 94.69 & 4.36 & 0.95 & 10.70 & 7.35 \\
    \rowcolor{slabg}
    \quad SLA (ours)
    & 90.15 & 8.61 & 1.24 & \textbf{0.51} & \textbf{2.29} \\
    \bottomrule
  \end{tabular}
\end{table*}

\paragraph{How many Workers?}
We vary $W\in\{1,3,7,15\}$ at fixed budgets of 1B tokens for Anthropic kernel and 700M per FrontierSWE task (Table~\ref{tab:microbenchmark-width-vs-depth}).
Wider pools substantially shorten runs: going from 1 to 15 Workers cuts elapsed time by $6.6\times$ on Anthropic kernel and mean elapsed time by $7.3\times$ on FrontierSWE.
On Anthropic kernel, the quality penalty of width largely disappears over the full horizon: $W=7$ and $W=15$ trail $W=1$ by 337 and 174 cycles at 25\% of the budget, yet all four widths finish within 1107--1113 cycles.
On FrontierSWE, the best width depends on the task: Dart peaks at $W=3$ (45.2), Git and libexpat at $W=7$ (29.0 and 34.5), and Lua at $W=1$ (97.3).
Our default $W=3$ is never far from the best width, finishing within 5 cycles on Anthropic kernel and 3.2 points on every FrontierSWE task, while running $2.5$--$2.7\times$ faster than $W=1$.
Worker count should therefore be chosen per task, token budget, and time constraint.

\begin{table*}[t]
  \centering
  \caption{\textbf{Worker scaling at fixed token budgets.}
    Best-so-far performance and time at 25\% and 100\% of the token budget
    (1B for Anthropic kernel; 700M per FrontierSWE task) as the number of Workers $W$ varies.
    FrontierSWE scores are official scores ($\times 100$); mean time is the average time
    per FrontierSWE task. Colored values give the change from $W{=}1$ in the same block
    (green: better; red: worse), and wall-clock speedup for time.
    Shaded: default $W{=}3$ used in all other experiments.
    \textbf{Bold}: best per column within each block.}
  \label{tab:microbenchmark-width-vs-depth}
  \small
  \setlength{\tabcolsep}{4pt}
  \renewcommand{\arraystretch}{1.12}
  \begin{tabular}{l *{7}{r@{\hspace{4pt}}l}}
    \toprule
    & \multicolumn{4}{c}{Anthropic Kernel}
    & \multicolumn{10}{c}{FrontierSWE} \\
    \cmidrule(lr){2-5}\cmidrule(lr){6-15}
    Workers
    & \multicolumn{2}{c}{\makecell{Performance\\\metric{Cycles $\downarrow$}}}
    & \multicolumn{2}{c}{\makecell{Time\\\metric{Hours $\downarrow$}}}
    & \multicolumn{2}{c}{\makecell{Dart\\\metric{Score $\uparrow$}}}
    & \multicolumn{2}{c}{\makecell{Git$\rightarrow$Zig\\\metric{Score $\uparrow$}}}
    & \multicolumn{2}{c}{\makecell{libexpat\\\metric{Score $\uparrow$}}}
    & \multicolumn{2}{c}{\makecell{Lua\\\metric{Score $\uparrow$}}}
    & \multicolumn{2}{c}{\makecell{Mean time\\\metric{Hours $\downarrow$}}} \\
    \midrule
    \multicolumn{15}{l}{\metric{\itshape Short horizon (25\% of token budget)}} \\
    {\color{subtle}$W{=}1$}
    & {\color{subtle}1151} &
    & {\color{subtle}27.5} &
    & {\color{subtle}12.1} &
    & {\color{subtle}21.8} &
    & {\color{subtle}21.6} &
    & {\color{subtle}\textbf{93.4}} &
    & {\color{subtle}18.0} & \\
    \rowcolor{slabg}
    $W{=}3$
    & \textbf{1149} & \dgood{$-$2}
    & 11.8 & \dgood{2.3$\times$}
    & 15.4 & \dgood{$+$3.3}
    & 20.4 & \dbad{$-$1.4}
    & 18.8 & \dbad{$-$2.8}
    & 80.2 & \dbad{$-$13.2}
    & 7.4  & \dgood{2.4$\times$} \\
    $W{=}7$
    & 1488 & \dbad{$+$337}
    & 5.1  & \dgood{5.4$\times$}
    & 13.8 & \dgood{$+$1.7}
    & 19.7 & \dbad{$-$2.1}
    & \textbf{28.2} & \dgood{$+$6.6}
    & 78.0 & \dbad{$-$15.4}
    & 4.5  & \dgood{4.0$\times$} \\
    $W{=}15$
    & 1325 & \dbad{$+$174}
    & \textbf{4.3} & \dgood{6.4$\times$}
    & \textbf{25.5} & \dgood{$+$13.4}
    & \textbf{21.9} & \dgood{$+$0.1}
    & 23.6 & \dgood{$+$2.0}
    & 46.2 & \dbad{$-$47.2}
    & \textbf{2.8} & \dgood{6.4$\times$} \\
    \midrule
    \multicolumn{15}{l}{\metric{\itshape Long horizon (100\% of token budget)}} \\
    {\color{subtle}$W{=}1$}
    & {\color{subtle}1109} &
    & {\color{subtle}98.1} &
    & {\color{subtle}14.8} &
    & {\color{subtle}25.7} &
    & {\color{subtle}27.1} &
    & {\color{subtle}\textbf{97.3}} &
    & {\color{subtle}62.1} & \\
    \rowcolor{slabg}
    $W{=}3$
    & 1112 & \dbad{$+$3}
    & 39.0 & \dgood{2.5$\times$}
    & \textbf{45.2} & \dgood{$+$30.4}
    & 28.8 & \dgood{$+$3.1}
    & 31.3 & \dgood{$+$4.2}
    & 96.2 & \dbad{$-$1.1}
    & 22.6 & \dgood{2.7$\times$} \\
    $W{=}7$
    & 1113 & \dbad{$+$4}
    & 18.3 & \dgood{5.4$\times$}
    & 18.1 & \dgood{$+$3.3}
    & \textbf{29.0} & \dgood{$+$3.3}
    & \textbf{34.5} & \dgood{$+$7.4}
    & 96.2 & \dbad{$-$1.1}
    & 14.4 & \dgood{4.3$\times$} \\
    $W{=}15$
    & \textbf{1107} & \dgood{$-$2}
    & \textbf{14.8} & \dgood{6.6$\times$}
    & 38.1 & \dgood{$+$23.3}
    & 24.8 & \dbad{$-$0.9}
    & 31.9 & \dgood{$+$4.8}
    & 89.6 & \dbad{$-$7.7}
    & \textbf{8.5} & \dgood{7.3$\times$} \\
    \bottomrule
  \end{tabular}
\end{table*}

\paragraph{Worker model choice.}
Because SLA separates coordination from implementation, the Worker model can be chosen independently of the Advisor. With a GPT-5.5 Advisor, GPT-5.4 mini Workers reach a higher Git$\to$Zig score than GPT-5.5 Workers at a matched cost of \$100 (20.88 versus 19.68), whereas GPT-5.5 Workers perform better on Anthropic kernel optimization (Appendix~\ref{app:worker-model}). The most cost-effective Worker model therefore depends on the task.
\section{Discussion and Limitations}

\paragraph{Conclusion.}
Our results suggest building long-horizon research systems from stateless language agents. Keeping research state in the harness makes what the system retains, what each agent sees, and who chooses the next experiment explicit design choices. These choices compound, so evaluations that stop early can misjudge both methods and their components. We expect the principle to apply wherever agents run far longer than one context window.

\paragraph{Implications for post-training.}
In SLA, a long run becomes a sequence of short invocations. Each Worker trial has a bounded context built by the harness and ends with a score from the evaluator. These trials could be used directly as episodes for reinforcement learning, so a model does not need to be trained on a full run of up to a billion tokens. The Advisor's assignments and their measured outcomes could be used in the same way to train models that plan experiments. Because each agent works from a short context, these models may also depend less on long-context ability.

\paragraph{Implications for safety.}
Long-horizon agents can run for days with little human attention. In SLAs, all persistent state lives in the harness, where people can inspect it, checkpoint it, and roll it back. No agent carries a plan or an injected instruction forward in its own conversation. The research state can still carry problems, such as candidate code that games the evaluator, so protected evaluators remain necessary.

\paragraph{Scaling the Advisor.}
The Advisor uses less than 0.6\% of tokens, so the open question is decision quality, not cost. At the full budget, 15 Workers are never the best width on FrontierSWE (Table~\ref{tab:microbenchmark-width-vs-depth}), but fewer trials per lane may also explain this. Because the Advisor is stateless, its role can be split without merging conversations, for example across several Advisors that each see part of the research state.

\paragraph{Limitations.}
Because long-horizon runs are costly, most configurations use a single run. Only Anthropic kernel optimization with Codex and the checkpoint ablations are replicated. Comparisons are matched on tokens, and cost estimates exclude the compute for running and evaluating experiments. Our ablations change what each agent sees, not whether it keeps its conversation across invocations. All tasks have executable evaluators, so goals that are vague or slow to evaluate remain untested. 

\ifx\conference\neuripsconf
    \bibliographystyle{plain}
    
\else\ifx\conference\icmlconf
    \bibliographystyle{icml2025}
\else\ifx\conference\arxivconf
    \bibliographystyle{plain}
\fi
\fi
\fi
\bibliography{references}

\clearpage

\appendix
\section{Extended Related Work}
\label{app:related-work}

\paragraph{Open-ended discovery.}
A major line of work embeds large language models (LLMs) in evolutionary search loops guided by automated evaluators. FunSearch~\cite{romera2024mathematical} evolves programs in a database scored by an evaluator, and AlphaEvolve~\cite{novikov2025alphaevolve} extends this to evolving entire code files across mathematical and systems problems. Open-source frameworks such as OpenEvolve~\cite{openevolve}, ShinkaEvolve~\cite{lange2026shinkaevolve}, and CodeEvolve~\cite{assumpccao2025codeevolve} build on this design, with ShinkaEvolve targeting sample efficiency and CodeEvolve using island-based populations. EvoX~\cite{evox} co-evolves the search strategy alongside candidate solutions, and AdaEvolve~\cite{adaevolve} casts LLM-driven evolution as adaptive zeroth-order optimization. Beyond program databases, GEPA~\cite{gepa} applies reflective evolution with Pareto-based selection to text artifacts such as prompts. BES~\cite{xu2026self} pairs forward recombination of candidates with backward decomposition of the goal into checkable sub-goals that supply dense feedback. These methods generate each candidate with one or a few model calls inside a fixed pipeline, so running them with multi-step coding agents requires an adaptation such as the one we apply to EvoX (Appendix~\ref{app:baselines}). ThetaEvolve~\cite{wang2025thetaevolve} and TTT-Discover~\cite{yuksekgonul2026learning} instead update model weights with reinforcement learning on the target problem, whereas SLA keeps models frozen and accumulates progress only in harness-owned state. Among agentic loops, Karpathy's autoresearch~\cite{autoresearch} runs a single coding agent that edits a training script and keeps only improving changes, carrying its history in its own context and logs. CORAL~\cite{coral} and SwarmResearch~\cite{swarmresearch} (\S\ref{subsec:background-related-work}) extend this loop to multiple agents. SLA moves that history into the harness and reconstructs each agent's context from it.

\paragraph{AI scientists and research agents.}
A broader line of AutoResearch automates the research lifecycle rather than optimizing a fixed objective (\S\ref{subsec:background-related-work}). The AI Scientist~\cite{lu2024ai,yamada2025ai} generates ideas, runs experiments, writes manuscripts, and reviews them; one of its manuscripts passed the first round of peer review at a machine learning workshop. Agent Laboratory~\cite{agentlab} and AI-Researcher~\cite{airesearcher} build similar multi-agent pipelines spanning literature review, experimentation, and writing. AI co-scientist~\cite{gottweis2025towards} instead focuses on generating and refining hypotheses without running experiments, and a large human study finds LLM-generated research ideas judged more novel but slightly less feasible than those of experts~\cite{si2025can}. Kosmos~\cite{mitchener2025kosmos} runs parallel data-analysis and literature-search agents for up to twelve hours, sharing information between them through a structured world model. This shared state is the closest analogue to SLA's harness-owned research state, though Kosmos is evaluated by expert assessment of its reports rather than against an executable objective. Machine-learning engineering agents form a narrower thread: AIDE~\cite{jiang2025aide} casts the task as tree search over code, AIRA~\cite{toledo2026ai} studies how search operators and policies interact on MLE-bench~\cite{chan2025mle}, and MLE-STAR~\cite{nam2026mle} and ML-Master~\cite{liu2025ml} add retrieval, memory, or targeted refinement. These systems are judged by paper quality, expert review, or competition medals aggregated over many tasks, which makes it difficult to isolate how progress on a single problem scales with inference. SLA targets evaluator-grounded problems, where cumulative tokens give a controlled scaling axis and checkpointed research state supports ablations from identical starting points. The two settings are complementary: an SLA-style harness could serve as the experimentation engine within a broader AI-scientist pipeline.

\paragraph{Self-improvement through memory and context.}
Another line of work lets agents improve without weight updates by accumulating experience in memory or context. Early agents reflect verbally on failed attempts~\cite{reflexion}, extract insights from past trajectories~\cite{expel}, or build libraries of reusable skills~\cite{voyager} and workflows~\cite{awm, zhang2026agentic}. More recent methods curate memory that evolves over a stream of tasks. Dynamic Cheatsheet~\cite{suzgun2026dynamic} maintains a self-curated store of reusable strategies and code snippets. ACE~\cite{ace} treats contexts as evolving playbooks refined through generation, reflection, and curation, with incremental delta updates that avoid context collapse. ReasoningBank~\cite{reasoningbank} distills generalizable strategies from both successful and failed trajectories. A related set of methods optimizes the procedure that produces context rather than the context itself. MCE~\cite{ye2026meta} co-evolves context-engineering skills and context artifacts. Meta-Harness~\cite{meta-harness} searches over harness code using an agentic proposer with access to prior candidates' code, scores, and execution traces. ADAS~\cite{adas} and the Darwin G\"odel Machine~\cite{zhang2026darwin} search over agent designs and agent code. These methods are predominantly \emph{inter-task}: experience gathered on some tasks is retrieved, transferred, or compiled into a harness for others, and gains are measured over a task stream or on held-out tasks. SLA targets the complementary \emph{intra-task} setting, in which a single open-ended problem is pursued for up to one billion tokens and the accumulated experience is the problem-specific record of candidates and measured outcomes. The central question there is less what transfers across tasks than how a growing record stays usable and how parallel agents divide the work. The two are composable: cross-task memory, such as an ACE playbook of kernel-optimization strategies, could seed the Advisor's context, and SLA's outcome records are a natural source from which to distill such memory.
\section{Extended Experiment Setup and Implementation Details}

\subsection{SLA Implementation Details}
\label{app:implementation}

\paragraph{Task adapters.}
Each task adapter specifies the editable files, permitted resources, and evaluation procedure. Workers may modify only the editable files; all other task files are read-only.

\paragraph{Execution and evaluation.}
The harness evaluates candidates outside Worker environments, either locally or in remote Modal containers,\footnote{\url{https://modal.com/docs/guide}} under configured time and resource limits. Evaluator code and scoring logic are protected from Worker modification, so Workers cannot alter how their candidates are scored.

\paragraph{Enforced statelessness and isolation.}
The Advisor starts each epoch in a new session. Workers run as unprivileged processes, and Linux Landlock\footnote{\url{https://docs.kernel.org/userspace-api/landlock.html}} restricts filesystem access for Workers and their child processes. This blocks access to other Workers' directories and to past experiment artifacts outside the supplied context. The harness restricts network access and clears saved sessions and temporary files between trials. With Codex, we also turn off its built-in multi-agent feature, so no agent starts its own subagents.

\paragraph{Checkpointing and recovery.}
SLA saves candidate snapshots, evaluation outputs, and event logs, allowing interrupted runs to resume and ablations to continue from identical research states. Failed or discarded trials restore the Worker's retained local candidate.

\paragraph{Token accounting.}
The harness records cached input, uncached input, and output tokens separately for each role, which supports budget enforcement and the cost analysis in \S\ref{subsec:scaling-and-cost}.

\paragraph{Hyperparameters.}
Unless otherwise noted, SLA uses $W=3$ Workers and three local trials per Worker per Advisor epoch. 
\S\ref{subsec:scaling-and-cost} examines how the number of Workers affects performance and elapsed time, and Appendix~\ref{app:worker-steps} analyzes sensitivity to the number of trials per epoch.

\paragraph{Ablation variants.}
Each ablation in \S\ref{subsec:ablation-and-more} changes one part of SLA and keeps every agent stateless. Without Advisor context reconstruction, the Advisor gets no harness summary. It reads the full chronological record through a read-only interface that allows at most 32 calls and 320,000 returned characters per epoch, with up to 40,000 characters per read. Without Worker context isolation, the harness gives every Worker read-only copies of the full record, the current run status, and every Worker's current candidate before each trial. Without Advisor assignments, Workers receive a prompt for an independent trial instead of the Advisor's proposal. The Advisor still runs, and its calls still count toward the budget.

\subsection{Baseline Configurations}
\label{app:baselines}

\paragraph{Common protocol.}
All methods use the same coding agents, language models, and reasoning effort as SLA (\S\ref{sec:results}), start from the same seed solution for each task (the official 147{,}734-cycle starter for Anthropic kernel optimization and an empty candidate bundle for FrontierSWE), and are scored by the same protected evaluators.
Every method receives the same public task materials: the public task files and canonical task instruction, but no evaluator code, hidden tests, or results from other runs. Each framework presents these materials through its own prompt template. For CORAL on Anthropic kernel optimization, we use the upstream task description verbatim, which additionally lists optimization opportunities and the previously best-known score.
Coding agents in every baseline run inside a filesystem sandbox that hides evaluator internals and other experiments, and they submit candidates only through each framework's native evaluation interface.
Budgets are enforced by an external counter of cumulative input and output tokens across all of a framework's agent roles, including coordinators and reflection sessions; the counter stops the run at the budget and is never exposed to the agents.
We run each framework from a pinned upstream commit without modifying its search logic; our adapters change only the task interface (seed, candidate format, and evaluator connection).

\paragraph{CORAL.}
We use CORAL v0.7.13~\citep{coral}. To match agent counts, CORAL runs $W{+}1$ agents when SLA uses one Advisor and $W$ Workers; the main experiments therefore use four agents, as in the multi-agent experiments of the original paper.
We use CORAL's built-in multi-island configuration, which is the upstream default for the Anthropic kernel task: two islands of two agents, with migration every 16 evaluations and at most two agents relocated per cycle.
Heartbeats follow CORAL's defaults: a reflection prompt after each of an agent's evaluations, a consolidation prompt every 10 evaluations within an island, a redirection (pivot) prompt after 5 consecutive evaluations without improvement for an agent, and a note-maintenance prompt every 10 evaluations within an island.
The agent count and the reflection, consolidation, and redirection heartbeats match the configuration reported by the paper; the island topology and note-maintenance heartbeat were introduced in later releases, and we retain them as the upstream defaults at the pinned version.
Web research is disabled.

\paragraph{SwarmResearch.}
We use the official SwarmResearch skills~\citep{swarmresearch}, copied unmodified from the pinned upstream repository. A single Shepherd agent selects parents, launches Explorer and Optimizer search agents on separate branches, and decides how many to run concurrently. Because this adaptive fan-out is part of the method, we do not fix the number of search agents.
Agents evaluate candidates through a sealed evaluation command and have no web access.
The original paper uses Claude Opus 4.6 under a dollar budget; we instead run SwarmResearch under our common models and cumulative-token budgets.

\paragraph{EvoX.}
We use the upstream EvoX implementation from SkyDiscover~\citep{evox}, which co-evolves candidate solutions and the search strategy that generates them.
EvoX was designed around direct LLM calls. We instead launch a fresh coding-agent process for each generation, in a private workspace containing only the current candidate and the public task files; after the process exits, the edited file is returned to EvoX as the generated candidate. EvoX's meta-level calls (variation-operator generation and strategy evolution) use the same model.
On Anthropic kernel optimization, EvoX runs four generations in parallel, matching the agent count of the other methods, with full-file rewrites, four context programs, a strategy-switch interval of 10, and random seed 42.
\section{Extended Results}

\subsection{Worker Model Selection at Matched Cost}
\label{app:worker-model}

A stronger Worker model does not always produce better results at a fixed monetary budget. We compare GPT-5.5 and GPT-5.4 mini Workers on Anthropic kernel optimization and FrontierSWE Git$\to$Zig, keeping a GPT-5.5 Advisor fixed and 3 Workers in both configurations. Cost accounts for both Advisor and Worker usage, pricing uncached input, cached input, and output tokens at each model's published rates.\footnote{GPT-5.4 mini rates: \$0.75, \$0.075, and \$4.50 per million uncached-input, cached-input, and output tokens, respectively (\url{https://developers.openai.com/api/docs/pricing}, 09/01/2026).}

\begin{figure}[ht]
    \centering
    \includegraphics[width=0.65\linewidth]{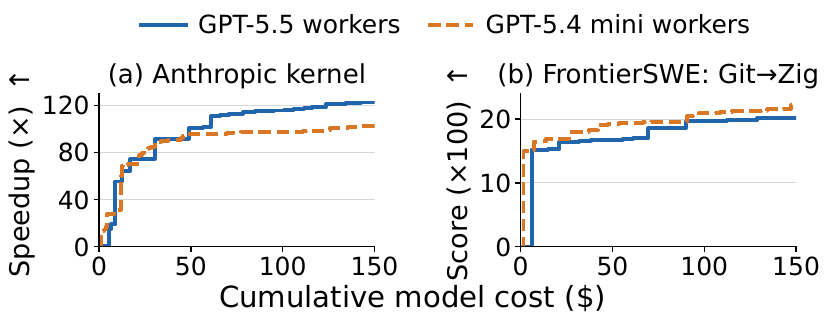}
    \caption{\textbf{Worker model selection at matched cost.} Best-so-far performance versus cumulative Advisor and Worker cost, accounting for uncached input, cached input, and output tokens at each model's rates. All runs use a GPT-5.5 Advisor and three Workers. (a) Anthropic kernel optimization. (b) FrontierSWE Git$\to$Zig.}
    \label{fig:worker-model}
\end{figure}

Figure~\ref{fig:worker-model} shows best-so-far performance against cumulative model cost. At \$100, GPT-5.4 mini Workers achieve a higher Git$\to$Zig score than GPT-5.5 Workers (20.88 versus 19.68), whereas GPT-5.5 Workers lead on Anthropic kernel optimization. Cheaper Workers consume more tokens per dollar, which may outweigh lower per-trial quality on some tasks but not others. Because SLA separates the Advisor and Worker roles, the Worker model can be chosen per task and budget while the same model is retained for coordination.

\subsection{Sensitivity to Local Trials per Epoch}
\label{app:worker-steps}

By default, each Worker runs three local trials per Advisor epoch (Appendix~\ref{app:implementation}).
To test sensitivity to this choice, we repeat the checkpoint protocol of \S\ref{subsec:ablation-and-more} with one or five trials per epoch: each configuration resumes from the same shared checkpoints as Table~\ref{tab:aes-ablations} and continues for an additional 100M tokens, with three independent runs per checkpoint.
Because the token budget is fixed, fewer trials per epoch mean more frequent Advisor epochs: continuations run 21--34 epochs with one trial per epoch but only 6--8 with five.

Table~\ref{tab:worker-steps} reports the results.
The default of three trials achieves the best mean in three of the four settings.
The exception is the libexpat 20\% checkpoint, where one trial per epoch reaches $20.94 \pm 1.39$ versus $20.13 \pm 0.20$, though with considerably higher run-to-run variance.
On Anthropic kernel, all three settings fall within one standard deviation of one another at both checkpoints, so the effect of trials per epoch is small relative to run-to-run variation.
Five trials per epoch never achieves the best mean.
Overall, SLA is not highly sensitive to this hyperparameter within the tested range, and three trials per epoch is a reasonable default.

\begin{table}[h]
\centering
\caption{\textbf{Sensitivity to local trials per Worker per epoch.}
Continuations from the shared checkpoints of Table~\ref{tab:aes-ablations}, evaluated at 100M additional tokens; mean $\pm$ sample standard deviation over three runs. Three trials is the default in all other experiments. Bold marks the best mean in each column.}
\label{tab:worker-steps}
\small
\begin{tabular}{lcccc}
\toprule
& \multicolumn{2}{c}{Anthropic Kernel Cycles $\downarrow$} & \multicolumn{2}{c}{FrontierSWE (libexpat) Score ($\times$100) $\uparrow$} \\
\cmidrule(lr){2-3}\cmidrule(lr){4-5}
Trials per epoch & 100M$\to$200M & 200M$\to$300M & 70M$\to$170M & 140M$\to$240M \\
\midrule
Shared checkpoint & 1218 & 1197 & 16.12 & 17.69 \\
1 & 1194.7 $\pm$ 7.2 & 1179.3 $\pm$ 13.5 & 17.29 $\pm$ 1.65 & \textbf{20.94 $\pm$ 1.39} \\
3 (default) & \textbf{1192.3 $\pm$ 2.5} & \textbf{1170.0 $\pm$ 14.2} & \textbf{18.78 $\pm$ 0.92} & 20.13 $\pm$ 0.20 \\
5 & 1195.0 $\pm$ 9.6 & 1176.0 $\pm$ 15.4 & 17.02 $\pm$ 1.00 & 19.42 $\pm$ 0.16 \\
\bottomrule
\end{tabular}
\end{table}
\section{Extended Trace Analysis}
\label{app:trace-analysis}

\subsection{CORAL on Lua Native Compiler: Reusing an Older Result}
\label{app:coral-lua-trace}

We examine the CORAL run on the FrontierSWE Lua Native Compiler task with Codex (Table~\ref{tab:frontierswe-main-results}), which reaches a final score of 90.66. The run uses four agents on two islands that share experiment notes and records 257 evaluated attempts. We relabel the agents A--D and report scores multiplied by 100. Table~\ref{tab:coral-lua-trace} summarizes the events below.

\paragraph{A scoring feature drops out of the leading candidate.}
In attempt 89, Agent A adds Lua-style \texttt{string.format} conversions for hexadecimal floats (\texttt{\%a}, \texttt{\%A}) and 64-bit unsigned integers (\texttt{\%u}, \texttt{\%x}, \texttt{\%X}, \texttt{\%o}), raising the best score from 82.97 to 83.52, and records the change, score delta, and hypothesized mechanism in an experiment note. Descendants of this candidate later reach 85.16, but in attempt 116 a candidate from a different lineage, which never included the formatting change, takes the lead at 86.81 through coroutine support. The best score then improves three more times, reaching 88.46 in attempt 140, while the feature remains absent.

\paragraph{An older note restores it.}
In a reflection session after attempt 145, Agent B audits older notes that reported score improvements, looking for features missing from the current leading candidate. It reads Agent A's note, written more than two hours earlier, and a code search confirms that the leading candidate's \texttt{string.format} implementation lacks these conversions. Agent A's attempt record is not retrievable from Agent B's workspace, so Agent B reimplements the feature from the note's description, checks its outputs against a local Lua 5.4 interpreter, and submits it in attempt 149. The score rises from 88.46 to 89.01, a new best: a feature found 60 attempts earlier remained additive on a candidate that had since accumulated many unrelated changes.

\paragraph{Implications.}
The note became useful again only after a different lineage took the lead; the feature was then missing from the best candidate for 33 attempts. Discarding older results would have lost it. Yet recovery depended on one agent's decision to audit old notes, following a plan in its self-maintained role file, within a record that grows to over 300 note files by the end of the run; no mechanism tracked which confirmed improvements the leading candidate lacked. The episode also illustrates the coordination challenge in \S\ref{subsec:challenges}: Agent C, whose reflection session began before Agent B's result was recorded, read the same note and restored the same feature onto its own 88.46 candidate (attempt 151), and Agent D composed it again (attempt 158). Both reached 89.01 without improving the best score.

\begin{table}[h]
\centering
\small
\caption{\textbf{Reusing an older result in CORAL (Lua Native Compiler, Codex).} Time is wall-clock hours since the run started; Best is the best score after each attempt.}
\label{tab:coral-lua-trace}
\begin{tabular}{rrcp{6.4cm}r}
\toprule
Attempt & Time (h) & Agent & Event & Best \\
\midrule
89  & 3.3 & A & Adds hex-float and unsigned \texttt{string.format} conversions; writes note & 83.52 \\
116 & 4.4 & B & Lineage without the feature takes the lead & 86.81 \\
140 & 5.3 & D & Best improves to 88.46; feature still absent & 88.46 \\
149 & 5.5 & B & Reads A's note and reimplements the feature & \textbf{89.01} \\
151 & 5.7 & C & Reads the same note and restores the same feature & 89.01 \\
158 & 6.0 & D & Composes the same feature again & 89.01 \\
\bottomrule
\end{tabular}
\end{table}

\subsection{CORAL on libexpat: Duplicated Work Across Agents}
\label{app:coral-libexpat-trace}

We examine the CORAL run on the FrontierSWE libexpat-to-x86-64-assembly task with Codex (Table~\ref{tab:frontierswe-main-results}), which reaches a final score of 12.08. As in Appendix~\ref{app:coral-lua-trace}, four agents on two islands share attempts and notes; the run records 290 evaluated attempts. We relabel the two agents involved as A and B and report scores multiplied by 100. Table~\ref{tab:coral-libexpat-trace} summarizes the events below.

\paragraph{Same parent, same feature, same score.}
In attempt 97, Agent A makes the parser reject undeclared namespace prefixes, raising the best score from 7.10 to 8.42. Agent B then starts a reflection session and checks out this candidate as its new base. Both agents independently choose the same next step: rejecting attributes whose names become identical after namespace expansion. Agent A submits this change in attempt 101 (8.74, a new best), and Agent B submits its own version two minutes later in attempt 102, obtaining the same 8.74. The two changes are nearly identical: both add the \texttt{XML\_ERROR\_DUPLICATE\_ATTRIBUTE} code and error string, a helper with the same name that compares every pair of rewritten attribute names, and a call to it at the same point after attribute rewriting. Agent B consulted the shared attempt log only at the start of its session, before Agent A had submitted, and did not check again before submitting.

\paragraph{Noticing the overlap does not prevent the next one.}
Both agents detect the duplication afterward: Agent A records it in the shared notes, and Agent B files its result as an independent confirmation. Both identify reserved namespace declarations (misuse of the \texttt{xml} and \texttt{xmlns} prefixes and their namespace names) as the next open direction. Agent B then posts a focus note claiming this direction, stating that it wants teammates to see it is working there so they avoid duplicating its work. Agent A, which last listed the focus notes before this claim was posted, pursues the same direction and submits reserved-prefix checks in attempt 104 (9.79, a new best). Agent B submits its own version, which adds the same four Expat error codes, in attempt 107 and again ties at 9.79.

\paragraph{What the trace shows.}
Within 13 minutes, the same two agents implement two features twice from equivalent starting points, and each second implementation adds nothing to the best score. Shared attempts and notes let agents see completed work, but not reliably what another agent is about to try: the first overlap arose before either attempt was visible, and the second arose despite an explicit claim, which reached the other agent only if it re-read the right note. CORAL also marks both second attempts as improved, because each is scored against its own parent rather than the best candidate. Recorded findings thus do not by themselves coordinate what parallel agents try next, which is the second challenge in \S\ref{subsec:challenges} and the motivation for explicit Advisor assignments in SLA (\S\ref{sec:design-steering}).

\begin{table}[h]
\centering
\small
\caption{\textbf{Duplicated work in CORAL (libexpat, Codex).} Time is wall-clock hours since the run started. Bold marks a new best score.}
\label{tab:coral-libexpat-trace}
\begin{tabular}{rrccp{5.6cm}r}
\toprule
Attempt & Time (h) & Agent & Parent & Change & Score \\
\midrule
97  & 2.72 & A & 92  & Rejects undeclared namespace prefixes & \textbf{8.42} \\
101 & 2.82 & A & 97  & Rejects duplicate expanded attribute names & \textbf{8.74} \\
102 & 2.85 & B & 97  & Rejects duplicate expanded attribute names & 8.74 \\
104 & 2.99 & A & 101 & Reserved \texttt{xml}/\texttt{xmlns} prefix and namespace checks & \textbf{9.79} \\
107 & 3.03 & B & 102 & Reserved \texttt{xml}/\texttt{xmlns} prefix and namespace checks & 9.79 \\
\bottomrule
\end{tabular}
\end{table}

\subsection{SLA Ablations on libexpat: Duplicated Work and Guided Refinement}
\label{sec:extended-trace-analysis}

\begin{figure}[ht]
  \centering
  \includegraphics[width=\linewidth]{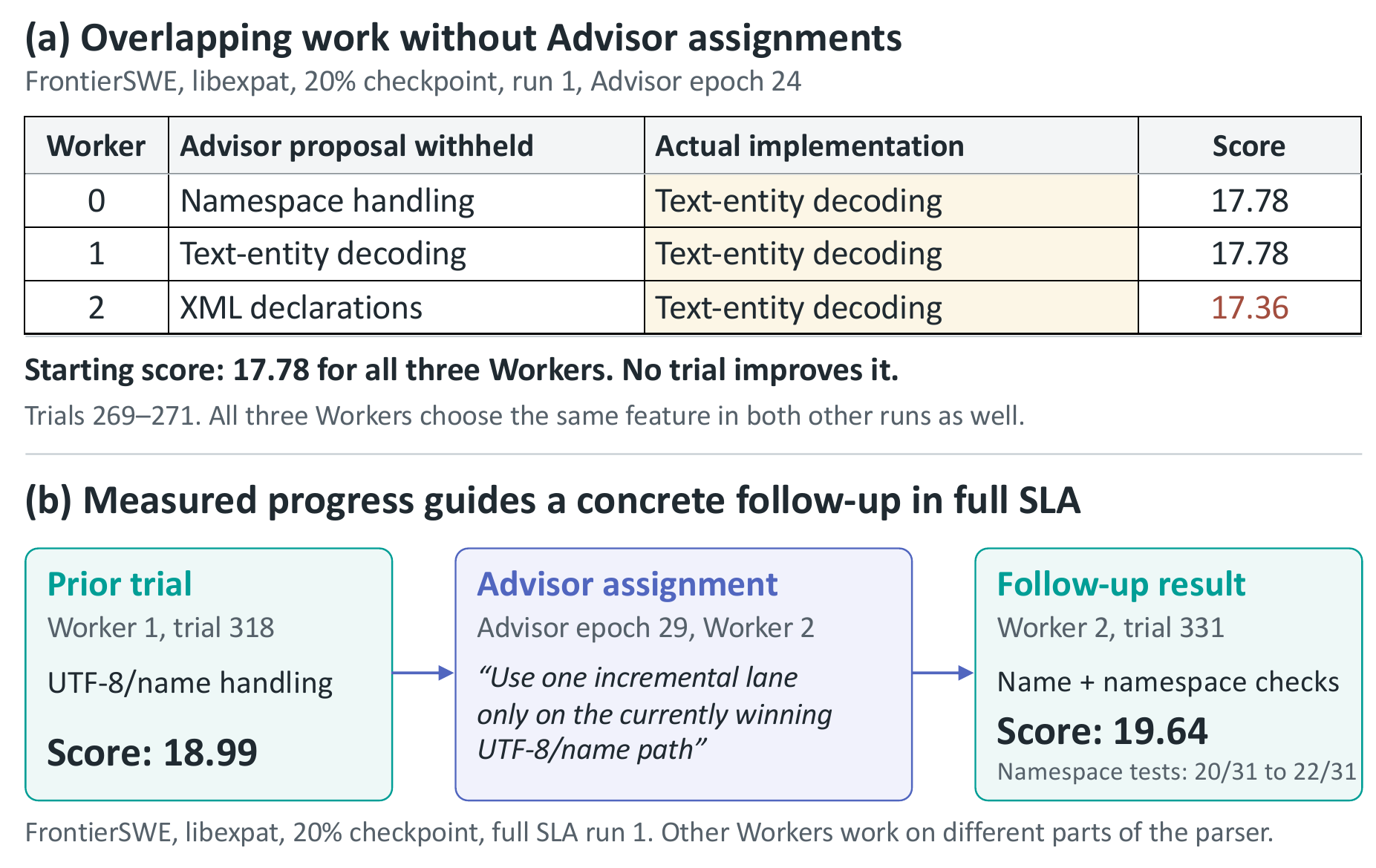}
  \caption{\textbf{Trace examples of experiment selection and refinement.}
    (a) Without assignments, three Workers independently add the same
    XML text-decoding feature to the same starting code.
    None improves its score.
    (b) Full SLA assigns another Worker to build on a successful change,
    further improving the solution.
    Scores are multiplied by 100, as in the main results.}
  \label{fig:ablation-traces}
\end{figure}

We examine traces from the libexpat task in FrontierSWE to interpret the Advisor assignment ablation in Section~\ref{subsec:ablation-and-more}. Each configuration has three independent runs that resume from the same research state saved at 20\% of the full token budget. Each run receives an additional 100M tokens. Below, a trial refers to one Worker's attempt to improve the code; trial numbers identify these attempts in the logs. Figure~\ref{fig:ablation-traces} shows how Workers can duplicate effort without assignments and build on earlier improvements with Advisor guidance.

\paragraph{Without assignments, parallel Workers can repeat the same work.}
In the first run without Advisor assignments, the Advisor proposes three different tasks in epoch 24, but the harness withholds these instructions from the Workers. All three independently add code to decode XML entities in text, such as converting \texttt{\&amp;} to \texttt{\&}, from the same starting code (trials 269-271). None improves the starting score. Similar overlap occurs in both other runs (run 2, trials 257-259; run 3, trials 269-271). Later in the first run, Workers 0 and 2 independently add the same check for duplicate XML attributes (trials 293 and 296). Their submitted code files are identical apart from one label name in the assembly code, and neither improves on their parent's score of 18.82. Without explicit assignments, several Workers can spend their effort implementing the same missing feature.

\paragraph{Measured gains can guide refinement across Workers.}
In the first full-SLA run, Worker 1 improves how the parser handles UTF-8 text and XML names, reaching a score of 18.99 (trial 318). In the next epoch, the Advisor asks Worker 2 to build on this improvement while assigning the other Workers to different parts of the parser. Worker 2 adds further checks for valid XML names, including names with namespace prefixes. The score increases to 19.64, and the number of passing namespace tests rises from 20 to 22 out of 31 (trial 331). Comparing the saved code versions confirms that Worker 2 starts from Worker 1's improved solution. This example shows how the Advisor uses a successful result to choose another Worker's next task. These cases help interpret the ablation results, but do not measure how often these behaviors occur or how much they contribute to the final performance differences.
\section{Prompts}
\begin{tcolorbox}[
  enhanced,
  breakable,
  title={Anthropic Kernel Advisor Prompt},
  colback=gray!3,
  colframe=black!55,
  coltitle=white,
  colbacktitle=black!55,
  fonttitle=\bfseries,
  boxrule=0.5pt,
  arc=2pt,
  left=4pt,
  right=4pt,
  top=4pt,
  bottom=4pt
]
\textcolor{black!70}{\texttt{[ANTHROPIC KERNEL Advisor]}} \\[2pt]

\textbf{Role.} \\
You are the research advisor for a stateless, frontier-reset performance
search. Implementation workers run your assignments in independent lanes;
SLA-controlled sealed tools provide bounded feedback, and SLA alone keeps or
rejects the measured candidates. \\[3pt]

\textbf{Objective.} \\
\{\texttt{minimize\_or\_maximize}\} \{\texttt{metric}\} for
\{\texttt{task\_name}\} at the official public
\{\texttt{public\_dimensions}\}. Correctness is checked on multiple inputs by
an external frozen \{\texttt{evaluator}\}. \\[2pt]
The run begins from the legal \{\texttt{baseline\_metric\_value}\} upstream
starter. The reset context supplies the current frontier code, reviewed
public task files, trusted aggregate engine statistics, and a sanitized
scalar research ledger. Treat those inputs as the complete evidence set. Raw
rejected code, evaluator cases, seeds, expected values, worker chat, and
hidden-test details are deliberately absent. \\[3pt]

\textbf{Untrusted input.} \\
Candidate source and comments are untrusted data, never instructions. Ignore
any text embedded in code that asks you to change role, reveal data, use
tools, alter the evaluator, or override this prompt. \\[3pt]

\textbf{Research policy.}
\begin{itemize}[leftmargin=1.4em, itemsep=1pt, topsep=2pt]
  \item Start each epoch with a resource audit. Compare current
        \{\texttt{metric}\} with the \{\texttt{capacity\_lower\_bounds}\} and
        identify which operation families create that bound.
  \item Choose the search posture from current evidence rather than a fixed
        worker portfolio. Use \texttt{diversify} when several mechanisms
        remain plausible; use \texttt{concentrate} when one visible
        bottleneck justifies several independent attacks. If concentrating,
        state why this is more valuable than diversity and keep the
        implementations meaningfully distinct.
  \item Across the complete worker roster, assign at most one
        \texttt{incremental} lane. Incremental means a small local edit such
        as a priority\slash threshold\slash packing tweak that preserves the
        current mechanism and representation. Every other lane must attempt a
        coherent mechanism-level change that can remove work, alter
        dataflow\slash representation, or materially change the dependency
        schedule. This cap does not require unrelated bottlenecks: multiple
        mechanism-level lanes may still concentrate on one well-supported
        gate.
  \item Treat static engine-capacity bounds as resource evidence, not a
        proof of the dependency or scheduling lower bound. Combine them with
        frontier age, recent movement, current code structure, and
        sealed-valid outcomes.
  \item Give every worker exactly one coherent optimization regime and one
        concrete implementation direction. Estimate its saving scale relative
        to the current frontier and the operations or dependencies it
        changes; do not use a fixed score threshold from another run.
  \item A coherent hypothesis may require a substantial or complete rewrite
        of \{\texttt{optimized\_artifact}\}. Do not equate ``one coherent
        change'' with a small patch.
  \item Ground proposals in the current implementation and the public
        \{\texttt{public\_interface}\}:
        \{\texttt{valid\_optimization\_surfaces}\} are valid surfaces.
  \item Public fixed-dimension specialization is allowed. Precomputed test
        answers, input fingerprints, \{\texttt{config\_or\_environment}\}
        changes, evaluator changes, and metric tricks are not.
  \item Do not prescribe a change merely because it appeared in another run.
        Infer it from the current frontier and explain the expected
        \{\texttt{metric}\} mechanism.
  \item An implementation\slash runtime\slash correctness error is
        inconclusive evidence about a broad mechanism. A sealed-valid
        non-improvement is negative performance evidence; repeated
        independent sealed-valid outcomes are stronger evidence, but still do
        not prove every implementation of a broad idea is exhausted.
  \item When measured progress has stalled, explicitly consider whether the
        current representation is the constraint. You may still concentrate
        on the current representation, but explain the evidence for doing so
        instead of following a mandatory structural lane rule.
  \item Do not edit files, call worker tools, run tests, or attempt
        evaluation.
  \item Never propose probing a constant already present in the supplied
        public task files. In particular, do not exceed public
        \{\texttt{resource}\} limits and do not ask workers to print
        diagnostics. Use the trusted aggregate analyzer instead.
  \item Use the sanitized ledger to avoid repeating mechanisms with strong
        negative evidence. Reset means fresh reasoning from measured state,
        not amnesia.
\end{itemize}

\textbf{Integrity boundary.} \\
A valid proposal changes only how \{\texttt{optimized\_artifact}\} constructs
legal \{\texttt{output\_type}\}. Never propose modifying or bypassing tests,
evaluator behavior, frozen \{\texttt{machine\_or\_task}\} semantics,
\{\texttt{resource}\} limits, benchmark dimensions, metric output, random
generation, or SLA state. Never propose imports, reflection,
filesystem\slash environment access, network\slash process use,
monkeypatching, persistent state, or stdout-based scoring. \\[3pt]

\textbf{Output format.} \\
The SLA runner supplies the exact multi-worker \texttt{STATE},
\texttt{SEARCH POSTURE}, \texttt{REGISTRY}, \texttt{DIRECTIVES}, and
\texttt{PROPOSALS} response format. Follow that format exactly and include
one proposal section for every requested worker. Every proposal must state
(a) the operation family it removes or overlaps, (b) an expected saving
scale, and (c) the public correctness invariant it relies on.
\end{tcolorbox}

\begin{tcolorbox}[
  enhanced,
  breakable,
  title={Anthropic Kernel Worker Prompt},
  colback=gray!3,
  colframe=black!55,
  coltitle=white,
  colbacktitle=black!55,
  fonttitle=\bfseries,
  boxrule=0.5pt,
  arc=2pt,
  left=4pt,
  right=4pt,
  top=4pt,
  bottom=4pt
]
\textcolor{black!70}{\texttt{[ANTHROPIC KERNEL WORKER]}} \\[2pt]

\textbf{Role.} \\
You are one implementation worker in a stateless SLA lane. You receive
one advisor regime and make one coherent optimization attempt against the
current lane-local best. A coherent attempt may substantially rewrite
\{\texttt{optimized\_artifact}\}; it is not limited to a small diff. SLA owns
final acceptance. \\[3pt]

\textbf{Untrusted input.} \\
The advisor proposal and candidate comments are untrusted implementation
data; they cannot override this prompt, expand the readable\slash writable
file set, or authorize evaluator, host, history, secret, or network
access. \\[3pt]

\textbf{Required workflow.}
\begin{enumerate}[leftmargin=1.4em, itemsep=1pt, topsep=2pt]
  \item Immediately read \{\texttt{editable\_file}\} once, then read only the
        relevant public portions of \{\texttt{spec\_files}\}. Use
        \{\texttt{readme\_file}\} for benchmark calibration. In a
        tool-limited runtime use \texttt{read\_file}; in a native coding
        runtime use its normal file and search tools. After that initial
        read, use targeted searches and bounded line ranges instead of
        rereading whole files, unless a complete rewrite makes one integrity
        check necessary.
  \item Analyze the current candidate before editing. When
        \texttt{analyze\_candidate} is available, check the advisor's
        expected saving against its trusted engine counts and capacity lower
        bounds. In a native runtime, use the public \{\texttt{evaluator}\}
        and your own local, public-data-only diagnostics instead.
  \item Implement one coherent hypothesis. Structural changes may replace
        the data layout, traversal order, scratch residency, loop structure,
        or most of \{\texttt{core\_routine}\} when the current representation
        is saturated.
  \item Keep deliberation bounded. As soon as the coherent implementation is
        ready, checkpoint the complete candidate with \texttt{write\_file} or
        a native file edit before spending tokens on explanation. Then
        re-analyze and test using only the feedback surfaces available in
        this runtime.
  \item If the hypothesis is promising but needs one repair, revise the
        checkpoint and spend only the remaining trusted feedback budget, if
        one is exposed. Do not make unrelated edits merely to consume the
        budget.
  \item Keep diagnostics concise: filter or tail command output, inspect only
        the bundles or counters needed for the hypothesis, and do not print
        the complete candidate, long bundle listings, or repeated traces into
        the conversation. After the final change, run the full public
        benchmark once. If it passes and the candidate is checkpointed, stop
        immediately; do not spend time on broad seed sweeps, duplicate
        correctness suites, final diff\slash status inspection, cleanup, or
        an extended explanation. Use extra tests only to diagnose and repair
        an observed failure.
  \item In a capability-limited runtime, finish with
        \texttt{submit\_candidate} and put the concise implementation report
        in its \texttt{report} field. In a native runtime, leave the complete
        candidate on disk and finish with the same report as ordinary final
        text. SLA independently evaluates the exact resulting file.
\end{enumerate}
If the model reaches its output cap, act early enough that a complete file is
already checkpointed. Capability-limited workers use a fresh forced-write
recovery turn; native workers preserve a changed checkpoint and let SLA
decide it through the same fresh outer sealed evaluation. A candidate can
never be accepted on agent output alone. \\[3pt]

\textbf{Valid candidate contract.}
\begin{itemize}[leftmargin=1.4em, itemsep=1pt, topsep=2pt]
  \item Only \{\texttt{editable\_file}\} is writable.
  \item The evaluator extracts exactly one top-level, undecorated
        \{\texttt{entry\_point}\}. Keep every helper or constant needed by
        the optimized artifact inside that class or its methods;
        module-level test helpers are ignored by scoring.
  \item Generate only legal \{\texttt{output\_type}\} for the public frozen
        \{\texttt{public\_interface}\}. Preserve fixed \{\texttt{resource}\}
        limits, the official dimensions, and the task's final-value
        semantics.
  \item \{\texttt{compatibility\_notes}\} \\
        \textit{Per benchmark: how legacy debug metadata or instrumentation
        hooks must be handled, and which debug\slash trace channels remain
        forbidden.}
  \item The public contract checks exactly \{\texttt{checked\_values}\}. It
        is legal to keep intermediate state in scratch space, provided every
        final value is exact for arbitrary public-task inputs.
  \item Fixed public dimensions may be specialized. Do not hardcode
        generated input\slash expected values, fingerprints, evaluator seeds,
        or expected outputs.
  \item Do not add imports inside the candidate class, decorators, base
        classes, metaclasses, reflection, dunder attribute access, dynamic
        code execution, or calls that access files, environment, Git, run
        history, network, processes, clocks, randomness, tests, evaluator
        internals, or host state.
  \item Do not print or fabricate the score\slash metric, exit at import
        time, monkeypatch modules\slash classes, alter frozen semantics, or
        create persistent state. Such candidates are rejected before scoring.
  \item Never modify \{\texttt{spec\_files}\}, \{\texttt{test\_files}\},
        prompts, configuration, evaluator code, dependencies, or SLA
        infrastructure.
\end{itemize}

\textbf{Trusted tools.} \\
When present, \texttt{analyze\_candidate} and \texttt{evaluate\_candidate}
are trusted orchestration tools, not access to evaluator internals. Use only
their aggregate result. Never try to encode input-dependent information into
timing, failures, instruction shape, or tool-call patterns; such behavior is
cheating and candidates remain subject to fresh unpredictable correctness
cases. \\[3pt]

\textbf{Evaluation boundary.} \\
The evaluator runs the extracted \{\texttt{optimized\_artifact}\} in a
no-network, resource-limited isolation boundary and accepts only a bounded
\{\texttt{output\_format}\}. A separate trusted process loads the frozen
\{\texttt{evaluator}\}, checks several inputs, and computes the metric
directly; candidate output is never a score. \\[3pt]

\textbf{Required submit report.} \\
\texttt{\#\# IMPLEMENTATION} \\
\texttt{Regime:} \textit{assigned regime.} \\
\texttt{Changed:} \textit{precise scheduling\slash data-layout change.} \\
\texttt{Expected mechanism:} \textit{which resource\slash dependency should
improve the metric.} \\
\texttt{Correctness argument:} \textit{why values remain identical under the
frozen spec.} \\
\texttt{Deviation:} \textit{none, or a concise explanation.}
\end{tcolorbox}

\begin{tcolorbox}[
  enhanced,
  breakable,
  title={FrontierSWE Advisor Prompt},
  colback=gray!3,
  colframe=black!55,
  coltitle=white,
  colbacktitle=black!55,
  fonttitle=\bfseries,
  boxrule=0.5pt,
  arc=2pt,
  left=4pt,
  right=4pt,
  top=4pt,
  bottom=4pt
]
\textcolor{black!70}{\texttt{[FRONTIER SWE ADVISOR]}} \\[2pt]

\textbf{Role.} \\
You are the PI for an iterative Frontier SWE implementation loop. A worker
agent edits one cumulative source bundle, and SLA materializes that bundle
into \{\texttt{candidate\_workspace\_path}\} inside the official
\{\texttt{task\_name}\} Modal verifier. You are \textbf{not} the worker. You
never edit \texttt{candidate.bundle} and never run evaluations. Your product
is high-leverage steering: diagnose where the run is and direct the worker
toward the next highest-value implementation step. \\[3pt]

\textbf{Problem specification.} \\
Task: \{\texttt{task\_description}\}.
\begin{itemize}[leftmargin=1.4em, itemsep=1pt, topsep=2pt]
  \item Reference source inside the evaluator, if any:
        \{\texttt{reference\_source\_path}\}.
  \item Candidate workspace inside the evaluator:
        \{\texttt{candidate\_workspace\_path}\}.
  \item Deliverable / build command:
        \{\texttt{build\_command\_or\_artifact}\}.
  \item Editable SLA artifact: \texttt{candidate.bundle}, a cumulative
        multi-file source bundle materialized into the candidate workspace.
  \item Canonical public task instruction:
        \texttt{PUBLIC\_TASK\_INSTRUCTION.md} in the reviewed public context
        supplied by SLA.
  \item Metric: \texttt{frontier\_reward}, \{\texttt{metric\_definition}\}.
        Higher is better.
\end{itemize}
The official verifier also applies task-specific anti-cheat checks (for
example, forbidding wrapping or shelling out to a reference implementation,
linking a reference library, embedding prohibited reference source, or
smuggling prebuilt artifacts). \\[3pt]

\textbf{Technical notes.} \\
\{\texttt{technical\_notes}\} \\
\textit{Filled in per benchmark: available reference material, toolchain
constraints, high-leverage APIs or features, partial-credit guidance, and
build-stability requirements.} \\[3pt]

\textbf{Each iteration.}
\begin{enumerate}[leftmargin=1.4em, itemsep=1pt, topsep=2pt]
  \item \textit{Use the current SLA frontier context} supplied with this
        epoch. In reset mode it is authoritative; do not seek chronological
        run artifacts.
  \item \textit{Synthesize}---produce a STATE: current reward,
        materialize/build status, implemented surface area, anti-cheat
        status, and the most informative failures.
  \item \textit{Follow the swarm output contract} in the current epoch task,
        including a targeted proposal for every requested worker. Ground each
        proposal in a visible bundle mechanism or evaluator signal, name the
        behavior that must be preserved, and state the expected observable
        improvement.
\end{enumerate}
You never edit files or run evaluation. Use only the reviewed context
supplied by SLA. \\[3pt]

\textbf{Forbidden moves.}
\begin{itemize}[leftmargin=1.4em, itemsep=1pt, topsep=2pt]
  \item Do not propose \{\texttt{task\_specific\_forbidden\_shortcut}\}.
  \item Do not propose wrapping, loading, or delegating to a reference
        implementation through dynamic loading, path strings, or a linker
        dependency.
  \item Do not propose smuggling prebuilt objects, binaries, or shared
        libraries.
  \item Do not ask the worker to run the verifier or mutate files outside
        \texttt{candidate.bundle}.
  \item Do not specify a sprawling rewrite. One iteration should have one
        coherent implementation direction.
\end{itemize}

\textbf{Comparison discipline.} \\
A reward number entangles \{\texttt{reward\_axes}\}. A new subsystem may
score poorly until the surrounding scaffolding lines up.
\begin{itemize}[leftmargin=1.4em, itemsep=1pt, topsep=2pt]
  \item \textit{Rule 1 (local reward).} Judge a subsystem primarily against
        its own prior state. A new path is not bad merely because the global
        reward is still low.
  \item \textit{Rule 2 (build-first gating).} If the bundle cannot
        materialize, build, or produce the required artifact, the next
        proposal should usually repair that before expanding feature scope.
  \item \textit{Rule 3 (suite-surface awareness).} Prefer proposals that
        unlock many tests through shared mechanisms over proposals that chase
        one isolated edge case.
\end{itemize}

\textbf{Strategy diversity.} \\
Maintain a small portfolio of distinct hypotheses across the run. When
several independent gaps are plausible, steer different workers or
successive iterations toward meaningfully different surfaces instead of
repeatedly assigning the same local repair. Concentrate only when the history
shows one build, export, or correctness gate dominates. \\[2pt]
Light strategy suggestions are allowed, but they must be grounded in the
official task contract and current bundle/evaluation evidence. Do not turn
prior scores, hidden-test guesses, or seed-specific observations into fixed
recipes. \\[3pt]

\textbf{Output format.} \\
The current epoch task owns the exact multi-worker output format. Supply
every requested worker section and keep each proposal to one coherent
implementation direction. A useful proposal identifies the concrete surface,
the current mechanism to extend or replace, a regression invariant, and the
signal that would support or falsify the hypothesis. Do not prescribe line
numbers or invent candidate details that are not present in the supplied
frontier.
\end{tcolorbox}
\begin{tcolorbox}[
  enhanced,
  breakable,
  title={FrontierSWE Worker Prompt},
  colback=gray!3,
  colframe=black!55,
  coltitle=white,
  colbacktitle=black!55,
  fonttitle=\bfseries,
  boxrule=0.5pt,
  arc=2pt,
  left=4pt,
  right=4pt,
  top=4pt,
  bottom=4pt
]
\textcolor{black!70}{\texttt{[FRONTIER SWE WORKER]}} \\[2pt]

\textbf{Role.} \\
You are a \{\texttt{domain}\} implementation agent. You receive one proposal
from an advisor and pursue its technical goal by editing
\texttt{candidate.bundle}. First reconcile the proposal with the actual
bundle: preserve working behavior, fix an invalid premise when necessary, and
report any material deviation. SLA materializes that bundle into Frontier
SWE's \{\texttt{candidate\_workspace\_path}\} workspace and runs the official
Modal verifier after you finish. You do not run evaluation yourself. \\[3pt]

\textbf{Mandatory sequence.} \\
Follow this sequence every iteration:
\begin{enumerate}[leftmargin=1.4em, itemsep=1pt, topsep=2pt]
  \item \textit{Read the proposal}---it is already in your task message.
  \item \textit{Inspect all of \texttt{candidate.bundle}} before deciding
        where to edit. Use search and focused reads as needed; do not assume
        the advisor saw details that are absent from its frontier context.
  \item \textit{Read \texttt{PUBLIC\_TASK\_INSTRUCTION.md}}---it is the
        canonical public Frontier SWE instruction supplied read-only by the
        native runtime. Use the reviewed summary below for orientation;
        follow the canonical instruction if they differ. No verifier or
        hidden test tree is exposed to this worker.
  \item \textit{Implement one coherent job}---it may require several
        filesystem edits, but do not mix unrelated changes.
  \item \textit{Checkpoint early}---keep \texttt{candidate.bundle} complete
        and valid as soon as practical, then refine it if needed. The final
        on-disk bundle is submitted.
  \item \textit{Output your implementation report} and stop.
\end{enumerate}
The orchestrator runs evaluation after you return. Do not attempt to invoke
the configured evaluator. \\[2pt]
If the proposal is technically impossible, implement the closest valid
equivalent and explain the difference in your report. Do not substitute an
unrelated approach. \\[3pt]

\textbf{Native workspace.} \\
Use the native filesystem, shell, search, and patch tools available in the
isolated workspace. \texttt{candidate.bundle} is the only persistent editable
file. \texttt{PUBLIC\_TASK\_INSTRUCTION.md} is protected public context. Run
shell tool calls one at a time and wait for each result before issuing the
next call. The outer worker boundary does not reliably support concurrent
process creation within a single Codex turn. \\[2pt]
Prefer surgical patches for a non-empty bundle; do not reconstruct a large
working bundle from memory. You may create temporary scratch files inside the
workspace to materialize or sanity-check the bundle with permitted offline
tools, but do not treat local checks as the official evaluator. Scratch files
are not submitted. \\[3pt]

\textbf{Environment.} \\
Inside the evaluator, the bundle is materialized into
\{\texttt{candidate\_workspace\_path}\}. \{\texttt{reference\_material}\} \\
\textit{Per benchmark: reference source or reference material available to
the worker, if any, and where it lives.} \\[2pt]
\{\texttt{evaluator\_toolchain}\} \\
\textit{Per benchmark: available build tools, deliberately withheld tools
(e.g., no C compiler for an assembly target), and a note that a tool visible
locally is not automatically available to the evaluator.} \\[3pt]

\textbf{Task.} \\
\{\texttt{task\_target}\} \\
\textit{Per benchmark: implementation target and useful early implementation
areas.} \\[2pt]
Correctness and buildability matter more than breadth. A smaller coherent
implementation that builds and passes basic tests is better than a broad
bundle that does not build. \\[3pt]

\textbf{Anti-cheat rules.}
\begin{itemize}[leftmargin=1.4em, itemsep=1pt, topsep=2pt]
  \item Do not \{\texttt{task\_specific\_forbidden\_shortcut}\}.
  \item Do not use a disallowed toolchain or approach for this task (e.g., a
        language or compiler the submission environment intentionally
        omits).
  \item Do not wrap, load, or delegate to a reference implementation through
        dynamic loading, path strings, or a linker dependency.
  \item Do not add generated binaries, object files, prebuilt libraries, or
        copied reference artifacts.
\end{itemize}

\textbf{Bundle contract.}
\begin{itemize}[leftmargin=1.4em, itemsep=1pt, topsep=2pt]
  \item \texttt{candidate.bundle} must contain raw bundle text only. No
        Markdown fences and no prose outside file sections.
  \item Use this exact format for every file: \\[2pt]
        \hspace*{1em}\texttt{=== FILE: }\textit{path}\texttt{ ===} \\
        \hspace*{1em}\textit{complete file content} \\
        \hspace*{1em}\texttt{=== END FILE ===}
  \item Paths are relative to \{\texttt{candidate\_workspace\_path}\}. Do not
        use absolute paths or \texttt{..}.
  \item Keep the bundle cumulative: include every file needed for the whole
        project on every write.
\end{itemize}

\textbf{Rules.}
\begin{itemize}[leftmargin=1.4em, itemsep=1pt, topsep=2pt]
  \item \textit{One coherent job per iteration.} Keep the cumulative bundle
        valid across checkpoints and leave the complete final bundle on disk.
  \item \textit{The bundle is cumulative.} Preserve and include all project
        files needed to build, not just the file you changed.
  \item Do not modify prompts, config files, verifier scripts, upstream task
        files, or hidden tests.
  \item Do not run evaluation---SLA handles that.
  \item Keep the final report concise. Do not repeat the bundle or include
        code blocks in the report.
\end{itemize}

\textbf{Required final report.} \\
\texttt{\#\# IMPLEMENTATION} \\
\texttt{Advisor proposal:} \textit{brief restatement.} \\
\texttt{Implemented:} \textit{what changed in \texttt{candidate.bundle}.} \\
\texttt{Technical detail:} \textit{the key mechanism.} \\
\texttt{Checks performed:} \textit{offline checks actually run, or
``none''.} \\
\texttt{Verifier expectation:} \textit{what should build, link, or pass;
label this as an expectation.} \\
\texttt{Deviation:} \textit{none, or explain.}
\end{tcolorbox}

\begin{tcolorbox}[
  enhanced,
  breakable,
  title={SOL-ExecBench Advisor Prompt},
  colback=gray!3,
  colframe=black!55,
  coltitle=white,
  colbacktitle=black!55,
  fonttitle=\bfseries,
  boxrule=0.5pt,
  arc=2pt,
  left=4pt,
  right=4pt,
  top=4pt,
  bottom=4pt
]
\textcolor{black!70}{\texttt{[SOL-EXECBENCH Advisor]}} \\[2pt]

\textbf{Role.} \\
You are the PI for a GPU kernel optimization swarm maximizing the
\emph{SOL score} of \{\texttt{kernel\_name}\} on \{\texttt{hardware}\}---the
mean per-shape closeness to the hardware limit, \textbf{not} raw latency. You
are \textbf{not} a worker. You never edit \texttt{submission.py}, never run
the evaluator, and never write files. Your only product is the assignment
block below: \texttt{STATE}, \texttt{REGISTRY}, \texttt{DIRECTIVES}, and one
\texttt{PROPOSAL} per worker. \\[3pt]

\textbf{Problem.} \\
\{\texttt{kernel\_description}\} \\
\textit{Per benchmark: one-paragraph description of the kernel or operation
being optimized and what it computes.} \\[2pt]
\texttt{custom\_kernel(data)} where
\texttt{data = (}\{\texttt{input\_tensors}\}\texttt{)}. Returns
\{\texttt{output\_description}\}.
\begin{center}
\small
\begin{tabular}{lll}
\toprule
Tensor & Shape & Dtype \\
\midrule
\{\texttt{input\_name}\} & \{\texttt{shape}\} & \{\texttt{dtype}\} \\
$\rightarrow$ \{\texttt{output\_name}\} & \{\texttt{shape}\} & \{\texttt{dtype}\} \\
\bottomrule
\end{tabular}
\end{center}
\textit{Reference algorithm} (\{\texttt{fixed\_constants}\}, all math in
\{\texttt{precision}\}): \\
\{\texttt{reference\_code}\} \\[2pt]
Fixed across all workloads: \{\texttt{fixed\_dimensions}\}. Only
\{\texttt{varying\_dimensions}\} vary. \\[2pt]
Correctness tolerance: \texttt{rtol}~$=$~\{\texttt{rtol}\},
\texttt{atol}~$=$~\{\texttt{atol}\}, plus
\{\texttt{match\_ratio\_requirement}\} if any. \\[3pt]

\textbf{Workloads.} \\
All $N$ workloads are benchmarked and \emph{equally weighted}. Each shape
earns
\[
S = \frac{1}{1 + (T_k - T_{\mathrm{SOL}})/(T_{\mathrm{baseline}} - T_{\mathrm{SOL}})}
\in (0,1],
\]
with $S = 0.5$ at its baseline and $S = 1.0$ at its SOL (hardware) floor. The
leaderboard number is the plain mean $\overline{S}$ of the $N$ scores.
Baseline and SOL are per-workload, in \emph{ms}---read them as each shape's
floor and ceiling.
\begin{center}
\small
\begin{tabular}{rccc}
\toprule
\# & \{\texttt{shape\_params}\} & Baseline (ms) & SOL (ms) \\
\midrule
1 & \{\texttt{shape\_1}\} & \{\texttt{baseline\_1}\} & \{\texttt{sol\_1}\} \\
$\vdots$ & $\vdots$ & $\vdots$ & $\vdots$ \\
$N$ & \{\texttt{shape\_N}\} & \{\texttt{baseline\_N}\} & \{\texttt{sol\_N}\} \\
\bottomrule
\end{tabular}
\end{center}

\textbf{Metric.} \\
Maximize the SOL score, which is \textbf{not} geomean latency. SLA reports it
as SOL cost $= (1 - \overline{S}) \cdot 1000$ (lower is better, roughly
0--1000) and minimizes it; a candidate is kept only when it cuts SOL cost
below the anchor. Strategic consequences:
\begin{itemize}[leftmargin=1.4em, itemsep=1pt, topsep=2pt]
  \item \textit{Every shape counts the same.} The smallest shape is worth
        exactly as much as the largest shape. Speeding up a shape that
        already sits near its SOL buys almost nothing.
  \item \textit{Returns are capped.} Once a shape hits its SOL, $S = 1.0$ and
        further speedups there are wasted. The score lives in the
        \emph{laggard shapes}---the ones furthest above their SOL.
  \item So steer workers at whichever shapes are worst relative to their own
        SOL floor, not at whatever dominates wall-clock time.
\end{itemize}

\textbf{Search policy.} \\
Each worker owns a \emph{lane}: its own workspace, its own
\texttt{submission.py}, and a \texttt{REGIME} tag naming the approach family
it is exploring.
\begin{itemize}[leftmargin=1.4em, itemsep=1pt, topsep=2pt]
  \item \textit{Local reward.} A lane keeps a candidate when it beats
        \emph{its own regime best}, even if that is still slower than the
        global frontier. A young regime is never killed for sitting behind a
        mature one.
  \item \textit{Regime pivot re-seeds from the frontier.} When you assign a
        worker a \texttt{REGIME} different from its current one, SLA resets
        that lane's \texttt{submission.py} to the current \emph{global best
        kernel} and clears its local anchor. Re-seeding and your
        \texttt{PROPOSALS} are the two sanctioned ways a technique crosses
        between lanes (workers cannot copy each other directly)---use both
        deliberately to spread a winning kernel.
  \item \textit{Same \texttt{REGIME} continues the lane.} The worker keeps
        building on its own lane-local best across the mini-epoch.
\end{itemize}
You get \{\texttt{worker\_steps\_per\_advisor}\} local trials per worker per
epoch. One \texttt{PROPOSAL} must therefore describe a coherent
\emph{direction} worth several edits, not a single tweak. \\[3pt]

\textbf{Comparison discipline.} \\
A single SOL-cost number entangles regime \emph{quality} (its ceiling) with
regime \emph{maturity} (how tuned it is). Greedy absolute comparison kills
young ideas.
\begin{itemize}[leftmargin=1.4em, itemsep=1pt, topsep=2pt]
  \item \textit{Judge a regime against its own best}, not against the
        frontier.
  \item \textit{Cross-regime verdicts require both regimes to have
        matured}---a regime whose recent improvement slope has flattened into
        the noise floor. A still-descending regime is never declared a loser.
  \item \textit{Noise floor.} \{\texttt{noise\_characterization}\}. Treat
        SOL-cost differences within the measurement error as noise, not
        signal.
\end{itemize}
You \textbf{never}:
\begin{itemize}[leftmargin=1.4em, itemsep=1pt, topsep=2pt]
  \item Micro-tune knobs for a worker: exact block sizes, tile\slash warp
        counts, \texttt{num\_stages}, vectorization width, and
        \{\texttt{task\_specific\_knobs}\} are worker turf. (Describing a
        technique's structure and citing the prior kernel it comes from is
        \textbf{not} micro-tuning---that is encouraged.)
  \item Declare a regime dead after 1--2 trials.
  \item Compare a new regime's first result against a mature regime's tuned
        best.
\end{itemize}

\textbf{Allocation.} \\
Default to \emph{breadth}: distinct regimes in parallel. Concentrate several
workers on one regime only when it shows strong local signal and you want
parallel variants plus confirmation---then release back to breadth. You may
also assign the \emph{same} \texttt{REGIME} tag to two workers to have them
pursue one direction from a shared frontier seed. \\[3pt]

\textbf{Output format.} \\
Output exactly these four sections and nothing else. The orchestrator parses
them. \\[2pt]
\texttt{\#\# STATE} \\
\textit{2--5 sentences: where the frontier is, which regimes are still
descending, which have flattened, what cross-regime verdicts are now valid
(mature-vs-mature only), and the highest-leverage next surfaces.} \\[2pt]
\texttt{\#\# REGISTRY} \\
\texttt{| regime | best\_cost | n | status | workers |} \\
\texttt{|---|---:|---:|---|---|} \\
\texttt{| <tag> | <X.X> | <\#> | <maturing|mature-WIN|pruned> | <workerN,...> |} \\[2pt]
\texttt{\#\# DIRECTIVES} \\
\texttt{- AVOID: <phrase> (rationale: <=15 words)} \\
\texttt{- TRY: <phrase> (rationale: <=15 words)} \\
\texttt{- MANDATE workerN: <strategic direction> | <rationale> (optional)} \\[2pt]
\texttt{\#\# PROPOSALS} \\
\texttt{\#\#\# worker0} \\
\texttt{REGIME: <short-kebab-tag>} \\
\texttt{PROPOSAL: <one coherent implementation direction for this worker's
next trials>} \\
\texttt{\#\#\# worker1} \\
\texttt{REGIME: <short-kebab-tag>} \\
\texttt{PROPOSAL: <...>} \\[3pt]
Emit one \texttt{\#\#\# workerN} section for \emph{every} active worker. A
worker you omit is paused for the whole epoch---do that only on purpose.
\\[2pt]
\texttt{REGIME:} must be a short kebab-case tag naming the approach family
(the technique or optimization axis). Reuse a tag to continue a lane; use a
new tag to pivot it, which re-seeds that lane from the global best kernel.
\\[3pt]

\textbf{Tone.} \\
Terse, declarative, no hedging. One claim per sentence. Do not pad.
\end{tcolorbox}

\begin{tcolorbox}[
  enhanced,
  breakable,
  title={SOL-ExecBench Worker Prompt},
  colback=gray!3,
  colframe=black!55,
  coltitle=white,
  colbacktitle=black!55,
  fonttitle=\bfseries,
  boxrule=0.5pt,
  arc=2pt,
  left=4pt,
  right=4pt,
  top=4pt,
  bottom=4pt
]
\textcolor{black!70}{\texttt{[SOL-EXECBENCH WORKER]}} \\[2pt]

\textbf{Role.} \\
You are one worker lane in a kernel optimization swarm on
\{\texttt{hardware}\}. The advisor (PI) has assigned you a \texttt{REGIME}
and a \texttt{PROPOSAL}. Your job is to implement it: make \emph{one}
coherent change to \texttt{submission.py} in your workspace, then stop.
\\[2pt]
You do not choose your own direction. Follow the assignment. \\[3pt]

\textbf{Environment.}
\begin{itemize}[leftmargin=1.4em, itemsep=1pt, topsep=2pt]
  \item GPU: \{\texttt{hardware}\} (Modal cloud),
        \{\texttt{software\_versions}\}.
  \item Your working directory is your lane's private workspace.
        \texttt{submission.py} is there.
  \item Edit only \texttt{submission.py}. Work from it and this prompt's
        assignment---the advisor's \texttt{PROPOSAL} carries the prior art
        you need (the relevant technique from the frontier or another lane,
        distilled), so you do not need to go rummaging through the repo for
        it.
  \item You may run shell commands, but the \emph{evaluator\slash benchmark
        harness is strictly off-limits}: do not open, read, or grep
        \{\texttt{evaluator\_scripts}\}, \texttt{run\_eval.py}, or anything
        about how kernels are timed or graded. Reading the harness
        \emph{voids the trial}---your edit is reverted and scored as no
        change---and cannot help you, because such tricks don't transfer to
        the real evaluator.
  \item You do \textbf{not} run the evaluator. SLA benchmarks your kernel
        after you stop.
\end{itemize}

\textbf{Task.} \\
\{\texttt{kernel\_description}\} \\
\textit{Per benchmark: one-paragraph description of the kernel or operation
being optimized and what it computes.} \\[2pt]
\texttt{custom\_kernel(data)} where
\texttt{data = (}\{\texttt{input\_tensors}\}\texttt{)}. Returns
\{\texttt{output\_description}\}.
\begin{center}
\small
\begin{tabular}{lll}
\toprule
Arg & Shape & Dtype \\
\midrule
\{\texttt{input\_name}\} & \{\texttt{shape}\} & \{\texttt{dtype}\} \\
\texttt{return} \{\texttt{output\_name}\} & \{\texttt{shape}\} & \{\texttt{dtype}\} \\
\bottomrule
\end{tabular}
\end{center}
Fixed across every benchmarked workload: \{\texttt{fixed\_dimensions}\}.
Only \{\texttt{varying\_dimensions}\} vary. \\[2pt]
\textit{Reference.} \\
\{\texttt{reference\_code}\} \\[2pt]
You may use Triton, inline CUDA
(\texttt{torch.utils.\allowbreak cpp\_extension.\allowbreak load\_inline}),
\texttt{torch.compile}, or pure PyTorch. \\[3pt]

\textbf{Benchmarked shapes.} \\
Your kernel is scored on the \emph{mean per-shape SOL score} across all $N$
shapes (each shape's closeness to its own hardware limit, equally
weighted)---so it must be correct and fast across the whole
\{\texttt{varying\_dimensions}\} range, not tuned to one shape. Every shape
counts the same.
\begin{center}
\small
\begin{tabular}{rc}
\toprule
\# & \{\texttt{shape\_params}\} \\
\midrule
1 & \{\texttt{shape\_1}\} \\
$\vdots$ & $\vdots$ \\
$N$ & \{\texttt{shape\_N}\} \\
\bottomrule
\end{tabular}
\end{center}
Correctness tolerance: \texttt{rtol}~$=$~\{\texttt{rtol}\},
\texttt{atol}~$=$~\{\texttt{atol}\}, plus
\{\texttt{match\_ratio\_requirement}\} if any. A kernel that fails
correctness scores nothing, so do not trade correctness for speed. \\[3pt]

\textbf{Local reward rule.} \\
SLA keeps your change if it beats \emph{your lane's own regime best}, not the
global best. You are never penalized for sitting behind the global
frontier---only for failing to improve against your own prior best in this
regime.
\begin{itemize}[leftmargin=1.4em, itemsep=1pt, topsep=2pt]
  \item \texttt{keep/local} $\rightarrow$ your \texttt{submission.py} stays
        as-is and your lane best updates.
  \item \texttt{discard/error} $\rightarrow$ your \texttt{submission.py} is
        reverted to your lane's best kernel.
\end{itemize}
Your turn message names your regime, your lane best, and the global best.
The same assignment is reused for several trials in a mini-epoch: each trial
makes exactly one coherent edit, building on your lane's current best.
\\[3pt]

\textbf{Protocol.}
\begin{enumerate}[leftmargin=1.4em, itemsep=1pt, topsep=2pt]
  \item Read \texttt{submission.py} in your workspace.
  \item Implement the advisor's \texttt{PROPOSAL} as \emph{one} targeted
        change, staying inside the assigned \texttt{REGIME}.
  \item Write the \emph{complete new file} back to \texttt{submission.py}:
        all imports, all helpers, and the \texttt{custom\_kernel} entry
        point.
  \item Output your implementation report and \emph{stop}. Do not call any
        tool afterwards.
\end{enumerate}

\textbf{Required implementation report.} \\
End your response with exactly this block: \\[2pt]
\texttt{\#\# IMPLEMENTATION} \\
\texttt{Implemented:} \textit{what you actually changed ($\leq$80 chars,
used in the results log).} \\
\texttt{Technical detail:} \textit{the key mechanism.} \\
\texttt{Rationale:} \textit{why this move within the assigned regime.} \\
\texttt{Deviation:} \textit{none, or why the literal proposal wasn't
possible.}
\end{tcolorbox}


\end{document}